\documentclass[conference]{IEEEtran}
\IEEEoverridecommandlockouts
\usepackage{cite}
\usepackage{amsmath,amssymb,amsfonts}
\usepackage{algorithmic}
\usepackage{graphicx}
\usepackage{graphicx}
\usepackage{textcomp}
\usepackage{xcolor}
\usepackage{enumerate}
\usepackage{tocloft}
\usepackage{subcaption}
\usepackage{amsmath}
\usepackage{textcomp}
\usepackage{algorithm}
\usepackage{mathtools}

\usepackage{float}
\usepackage{subcaption}
\usepackage{array}
\usepackage{pdfpages}
\usepackage{comment}
\usepackage{xspace}
\usepackage{float}

\newcommand{\ddcrnn}{\texttt{DDCRNN}\xspace}

\begin{document}


\title{Dynamic Graph Neural Network for \\
Traffic Forecasting in Wide Area Networks}


\author{
\IEEEauthorblockN{Tanwi Mallick}\IEEEauthorblockA{Argonne National Laboratory\\Lemont, Illinois \\ tmallick@anl.gov}\and\IEEEauthorblockN{Mariam Kiran}\IEEEauthorblockA{Lawrence Berkeley\\ National Laboratory\\Berkeley, California\\mkiran@lbl.gov }\and\IEEEauthorblockN{Bashir Mohammed}\IEEEauthorblockA{Lawrence Berkeley\\ National Laboratory\\Berkeley, California\\bmohammed@lbl.gov}\and\IEEEauthorblockN{Prasanna Balaprakash}\IEEEauthorblockA{Argonne National Laboratory\\Lemont, Illinois \\pbalapra@anl.gov}
}



\maketitle

\begin{abstract}

Wide area networking infrastructures (WANs), particularly science and research WANs, are the backbone for moving large volumes of scientific data between experimental facilities and data centers. With demands growing at exponential rates, these networks are struggling to cope with large data volumes, real-time responses, and overall network performance. Network operators are increasingly looking for innovative ways to manage the limited underlying network resources. Forecasting network traffic is a critical capability for proactive resource management, congestion mitigation, and dedicated transfer provisioning.  
To this end, we propose a nonautoregressive graph-based neural network for multistep network traffic forecasting. 
Specifically, we develop a dynamic variant of diffusion convolutional recurrent neural networks to forecast traffic in research WANs. We evaluate the efficacy of our approach on real traffic from ESnet, the U.S. Department of Energy's dedicated science network. Our results show that compared to classical forecasting methods,  
our approach explicitly learns the dynamic nature of spatiotemporal traffic patterns, showing significant improvements in forecasting accuracy. Our technique can surpass existing statistical and deep learning approaches by achieving $\approx$20\% mean absolute percentage error for multiple hours of forecasts despite dynamic network traffic settings. 

\end{abstract}

\IEEEpeerreviewmaketitle


\section{Introduction}
Large scientific experimental facilities, with high-speed data production rates, are bringing enormous data movement and transfer challenges to the underlying network infrastructure that supports distributed science workflows. With this relentless growth, there is a need to develop proactive planning, resource allocation, and provisioning methods to manage the current backbone bandwidth of research wide area networks (R-WANs), while keeping costs low \cite{cisco}.
One of these proactive strategies could rely on network operators' ability to forecast future traffic on the network based on the current state of the network. For example, if we can forecast the traffic for several hours in the future, we can minimize the effect of congestion by diverting flows from congested links to free or less congested links; similarly, dedicated lines can be assigned to large transfers so that they do not interfere with other data transfers. The ability to forecast congestion patterns a few hours ahead can enable new traffic management strategies with traditional network protocol design, essentially allowing one to utilize underused bandwidth more efficiently and minimize overall congestion in the network infrastructure. 





Network traffic forecasting in R-WANs is a formidable task, owing to a lack of regular patterns in how users access and perform data transfers over the network \cite{Boutaba2018}. Compared to Internet WANs, which have periodic patterns \cite{claffyphd}, R-WANs have random traffic spikes that are difficult to understand and anticipate. The traffic on R-WANs depends on which science experiments and devices are running and which groups are involved, and it is characterized by high-variability data transfers lasting minutes or even hours \cite{8648789}. 

Network monitoring tools such as Simple Network Management Protocol (SNMP), sflow, and netflow \cite{6814316} allow collecting traffic information on network nodes and flow transfers as time-stamped data recording gigabytes of log files (GB). Most monitoring tools collect data at 30-sec intervals, giving a fine-grained view that includes other key features such as protocols used (e.g., TCP, UDP), interfaces, source and destination IP addresses, and even flow speeds. These data sets can be leveraged for developing data-driven forecasting methods. Classical statistical time-series forecasting methods such as ARIMA and Holt-Winters have been investigated for network traffic forecasting \cite{Yoo:2016:TFM:2989742.2989755}, but have been less effective for R-WANs, as regular and seasonal patterns do not exist \cite{cisco}. Forecasting methods based on classical ML methods such as random forest and SVM have been developed to provide per-site forecasts \cite{lau2008local}, but they fail to consider the whole network and its spatial patterns and connectivity, making them less robust with respect to the dynamic nature of the R-WAN traffic \cite{zhao2010wind}. Recent improvements in data collection present new opportunities for developing deep learning (DL) methods for forecasting R-WAN traffic \cite{wang_DLstacked}. For example, variants of convolutional neural networks (CNNs) and long-short-term memory methods (LSTMs) have been shown to provide better accuracy than classical statistical techniques for data sets with seasonality \cite{arimagarchmodel}. Nevertheless, existing DL methods for network forecasting are out-of-the-box methods and are not particularly customized for the R-WAN. 
They do not take into account the spatial correlation of the entire network and the dynamic temporal correlation of R-WANs. 


We model the R-WAN networking infrastructure as a graph, where nodes and edges correspond to sites and their connectivity, modeling network traffic as time series. We propose a nonautoregressive graph neural network approach to forecast traffic for multiple time steps ahead. Specifically, we develop a dynamic diffusion convolutional recurrent neural network (\ddcrnn) that considers data movement as a diffusion process from one node to another based on connectivity. The spatial and temporal correlations are learned through graph diffusion convolution and recurrent units, respectively. Consequently, a single model is trained and used to forecast traffic on the entire R-WAN network. 
The \ddcrnn that we propose is based on the diffusion convolutional recurrent neural network (DCRNN), which was originally developed for highway transportation forecasting \cite{li2017diffusion,mallick2019graph}. The key difference between the highway traffic DCRNN and our \ddcrnn is the way in which the connectivity between the nodes is considered. In the DCRNN, the connectivity is static and computed on the basis of (driving) distance; in \ddcrnn, the connectivity is dynamic and computed on the basis of the current state of the traffic in the network. This approach is designed to explicitly model the dynamic nature of the R-WAN traffic. Our main contributions are as follows:
\begin{itemize}
    \item We develop a dynamic diffusion graph recurrent neural network architecture to forecast traffic for multiple time steps on an R-WAN that is characterized by a lack of regular patterns and seasonality. 
    \item We demonstrate the effectiveness of the proposed method on real data from the Energy Sciences Network (ESnet), a high-performance R-WAN built by the U.S. Department of Energy (DOE) to support U.S. scientific research. We show that our approach explicitly learns the dynamic nature of spatiotemporal traffic patterns in this R-WAN and outperforms existing statistical and DL methods used for traffic forecasting by achieving $\approx$20\% mean absolute percentage error for multiple hours of forecasts. 
\end{itemize}

Using real traffic data from ESnet, our model captures key patterns among sites and links, highlighting several interesting behaviors and relationships that were not previously known, such as some sites having patterns that are more amenable for forecasting than others.  

\section{Related Work}


Wolski et al. \cite{Wolski:1999:NWS:334556.335068} developed a TCP performance network forecasting method to help schedule computations over distributed networks. These authors utilized statistical approaches such as ARIMA, Holt-Winters and Hidden Markov, and were successful in predicting a few time-steps by modeling traffic as stochastic processes \cite{armstrongbook}. Nevertheless, these methods are dependent on finding seasonality and using it to improve predictions. Network traffic spikes randomly with sudden data transfers and lacks seasonal patterns \cite{cope}, affecting the forecasting accuracy of these methods \cite{Yoo:2016:TFM:2989742.2989755}. Moreover, these approaches also fail to learn long-range dependency \cite{Ntlangu:2017:MNT:3175684.3175725}. 


Research in software-defined networking (SDN) promises to provide flexible solutions for building agile networks and leveraging active monitoring, prediction, and informed decision-making \cite{McKeown:2008:OEI:1355734.1355746}. Google \cite{b4} used SDN to optimize link usage by doing "what-if" scenarios to schedule transfers. Using Multi-Protocol Label Switching, advanced forwarding schemes can control and optimize flows for packet forwarding \cite{Pathak:2011:LIM:2068816.2068859}. However, the optimization techniques discussed do not exploit forecasting algorithms to make decisions.    

Most R-WAN network links are underutilized, especially as routing protocols use path-finding algorithms  rather than accounting for current traffic utilization. In prior work, dynamic congestion-based routing algorithms have been developed to adapt as traffic changes \cite{internettcptraffic}. However, these algorithms are susceptible to oscillatory behaviors and performance degradation, as shown in \cite{neutm}.

The European R\&E network, GEANT, used LSTM models to forecast traffic on European links, but only demonstrated 15-min forecasts \cite{neutm}. Similarly stacked autoencoders \cite{wang_DLstacked} were used for 15-, 30- and 60-minute forecasts. However, these authors did not use highly dynamic R-WANs such as ESnet and exploited relatively simpler and shorter forecasts. Recently, DCRNN implementations, without any modifications, were successfully applied to forecast congestion events in a WAN \cite{8845132}. Network traffic forecasting has explored statistical and simpler ML models as seen in  \cite{Yoo:2016:TFM:2989742.2989755}\cite{lau2008local} \cite{zhao2010wind}, but prediction accuracy is seen to suffer because of lack of seasonality in R-WAN traffic traces.

In this paper, we show that DCRNN alone is not effective, as it does not consider the dynamic spatial and temporal traffic patterns in R-WAN networks.

\section{U.S. Research Network: ESnet} 
\label{sec_dataset}

The DOE science network ESnet provides services to more than 50 research sites and universities, including supercomputing facilities and major scientific instruments. ESnet also connects to 140 research and commercial networks, permitting geography-free collaboration around the world.

\begin{figure*}[!ht]
  \centering
    \includegraphics[width=0.9\linewidth]{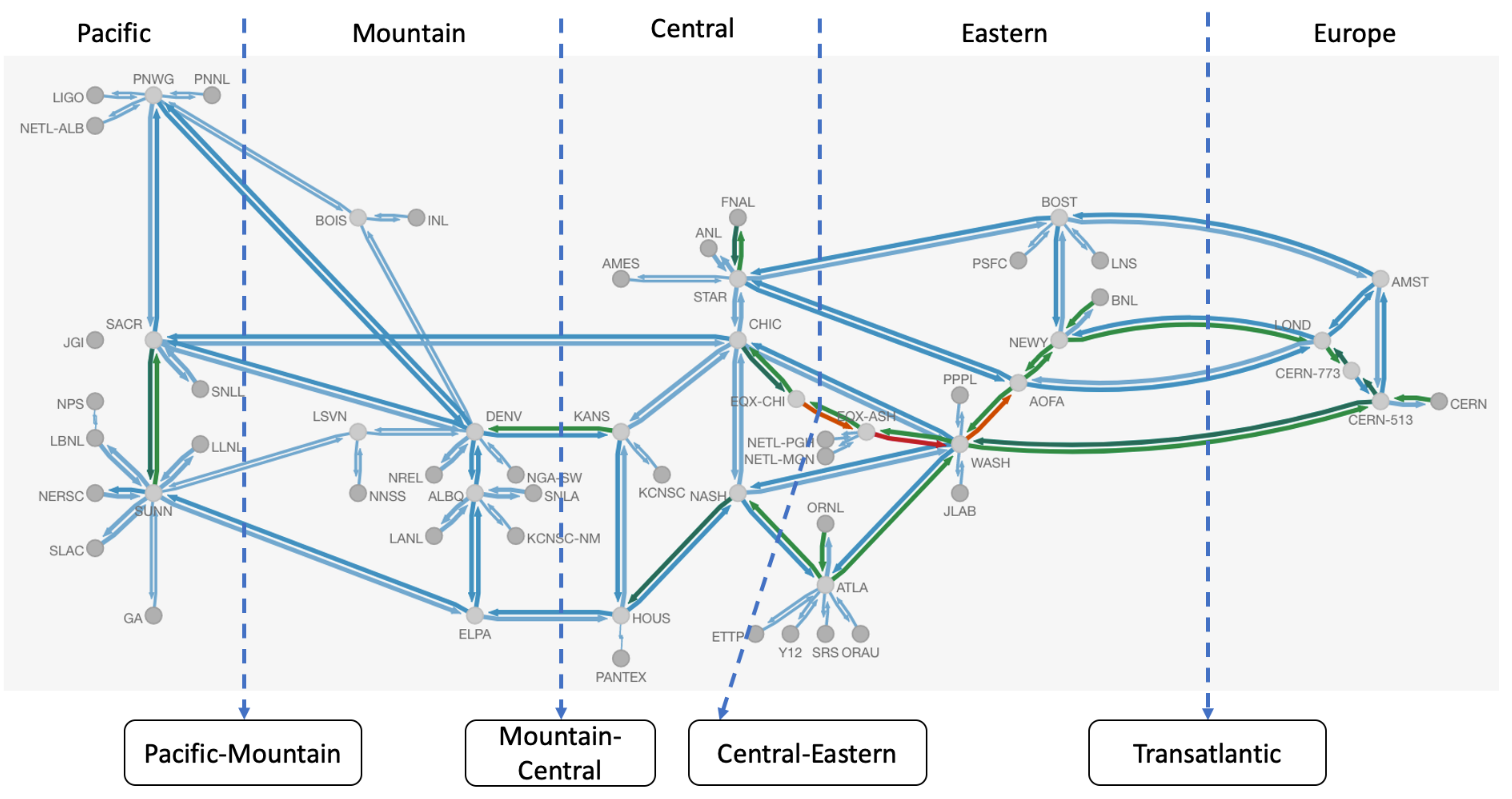}
  \caption{Full topology of ESnet.} 
  \label{fullgraph}
\end{figure*}

\begin{figure}[t]
  \centering
    \includegraphics[width=1\linewidth]{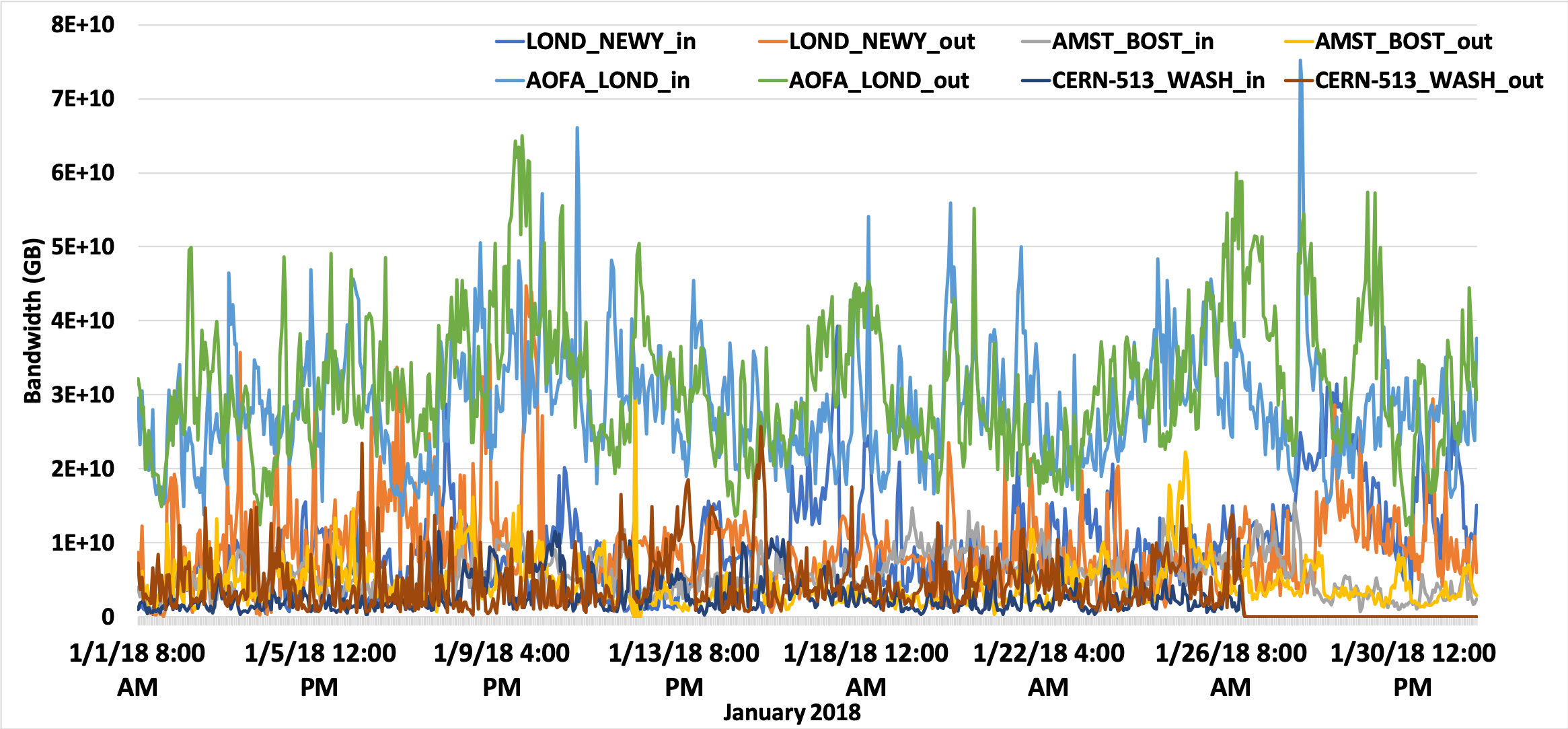}
  \caption{Traffic patterns in January 2018 on all transatlantic links in ESnet, including London, New York, Amsterdam, Boston, AOFA, CERN, and Washington, D.C. 
  }
  \label{fig:data}
\end{figure}
We use real traffic traces of two-way traffic data, collected from all of 2018, across the 48 sites in the U.S. and Europe. Figure \ref{fullgraph}
shows the full network topology of the network, divided among five time zones. All data are collected in GMT. Sample network traffic patterns on transatlantic links in ESnet appear in Figure \ref{fig:data}, which shows the lack of temporal patterns in the network traffic.

\subsection{SNMP Data Collection}

Network operators monitor link capacity and data movement across routers using SNMP \cite{feit1993snmp}.
Table \ref{snmpdata} shows a snapshot of these data during a two-way transfer between Sunnyvale and Sacramento in California.

The traffic is collected in moving GBs across router interfaces at 30-sec intervals. These data are then aggregated to 1-hour intervals and modeled as discrete observations. Aggregating traffic compounds the burst patterns \cite{internettcptraffic} and reduces the model complexity, allowing faster training and predictions. 

\begin{table}[!ht]
\centering
\begin{tabular}{|c|c|c|}
\hline
\centering
Timestamp     & SACR\_SUNN\_in (GB) & SACR\_SUNN\_out (GB)    \\ \hline
 	
1514822400 & 14110930202 &  1025131246 \\ \hline
1514826000 & 13453619303 &  9191557943 \\ \hline
 	
1514829600 & 12168879944 &  7793842045 \\ \hline
1514833200 & 11231198033 &  7097237528 \\ \hline
1514836800 & 10780847622 &  8048293939 \\  \hline
\end{tabular}\caption{Sample timestamped traffic trace collected from one router collecting traffic in both directions between Sacramento and Sunnyvale. 
}\label{snmpdata}
\end{table}

The data that we used for our study spans from 1 January 2018 to 31 December 2018, covering 48 sites with 96 traffic traces of two-way traffic.

While building hourly traffic summaries, we found that router interfaces sometimes miss recording traffic movements at specific intervals. To address this issue, we calculated the average data values for the missing points from the surrounding values to fill in the gaps. These missing values were only seen in the  Washington-Chicago link during a 1-week interval in November.

\section{Dynamic Diffusion Convolution Recurrent Neural Network}




We model the R-WAN as a graph   
$G = (V, E, A)$,  where $V$ is a set of $N$ nodes that represents the site; $E$ is a set of directed edges representing the connection between nodes, and $A \in R^{N\times N}$ is the weighted adjacency matrix representing the strength of connectivity between nodes.
Given the historical traffic observations at each node of the graph, the goal is to learn a function $\text{f(.)}$ that takes traffic observations for $T'$ time steps as input to forecast the traffic for the next $T$ time steps:
\begin{equation*}
X(t-T'+1), ...,X(t); G \xrightarrow{\text{f(.)}} X(t+1),... , X(t+T)
\end{equation*}

\begin{figure*}[t]
  \centering
    \includegraphics[width=1.0\linewidth]{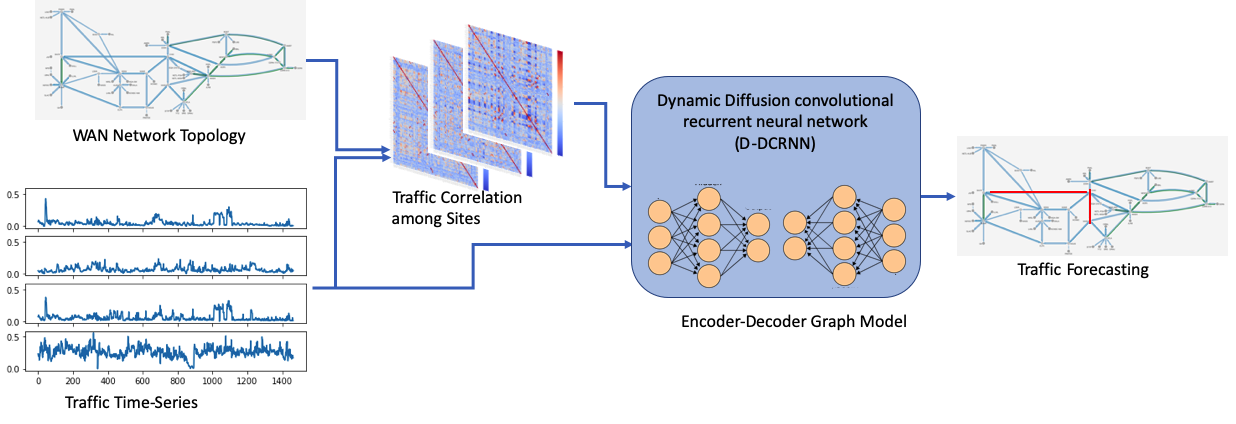}
  \caption{\ddcrnn model architecture. It takes an adjacency matrix computed from the current state of the network traffic among the sites of the WAN network topology and  the traffic as a time-series at each node of the graph. The encoder-decoder deep neural network is used to forecast the network traffic for mutiple time steps.}
  \label{model-arch}
\end{figure*}

\ddcrnn utilizes a graph-based encoder-decoder architecture that models spatial and temporal patterns through diffusion convolution operation on a graph and gated recurrent units (GRUs), respectively. Specifically, the matrix multiplication operations of the GRU cell are replaced by a diffusion convolution operation to perform graph convolution on the time-series data. A unit in the \ddcrnn  architecture is given by:
\begin{equation}\label{equ_1}
\begin{array}{lcl}
r^t & = &  \sigma (W_{r\bigstar G} [X_{t}, h_{t-1}] + b_r) \\
u^t & = &  \sigma (W_{u\bigstar G} [X_{t}, h_{t-1}] + b_u) \\
c^t & = &  \tanh (W_{c\bigstar G} [X_{t} (r_t \odot h_{t-1}] + b_c) \\
h_t & = & u_t \odot h_{t-1} + (1 - u_t) \odot c_t ,
\end{array}
\end{equation}
where $X_{t}$ and $h_t$ denote the input and final state
at time $t$, respectively; $r_t$, $u_t$, and $c_t$ are the reset gate, update gate, and cell state at time $t$, respectively; $\bigstar G$ denotes the diffusion convolution operation; and $W_r, W_u,$ and $W_c$ are parameters of the GRU cell. The diffusion convolution operation \cite{li2017diffusion} on the graph $G$ and the input data $X$ is defined as:
\begin{equation}
\label{equ_2}
 W_{\bigstar G} X = \sum_{d=0}^{K-1} (W_O (D_O^{-1}A)^d + W_{I}(D_I^{-1}A)^d) X,
\end{equation}
where $K$ is a maximum diffusion step;  $D_O^{-1}A$ and $D_I^{-1}A$ are transition matrices of the diffusion process and the reverse one, respectively; $D_O$ and $D_I$ are the in-degree and out-degree diagonal matrices, respectively, and $W_O$ and $W_I$ are the learnable filters for the bidirectional diffusion process.

To model the dynamic nature of the R-WAN traffic, \ddcrnn uses a dynamic weighted adjacency matrix estimated from the time-series data.
Specifically, instead of keeping $A$ as static, for each input sequence of $T'$ time steps on $N$ nodes (for both training and inference), \ddcrnn computes the linear correlation coefficient values between sites that have data transfer between them and uses it in Eq.~ \ref{equ_2}. Consequently, the elements in the adjacency matrix $\tilde{A}$ at time $T$ are given by: 
\begin{equation}
\tilde{A}_{ij} = \rho_{X_i, X_j} = \frac{cov(X_i, X_j)}{\sigma_{X_i}, \sigma_{X_j}},
\label{corr:equa}
\end{equation}
where, $A_{ij}$ represents the edge weight between the nodes $N_i$ and $N_j$; $X_i$ and $X_j$ are the time-series data of nodes $N_i$ and $N_j$; $\rho$ is the Pearson correlation coefficient between $X_i$ and $X_j$; $cov$ is the covariance;
$\sigma _{X_i}$ is the standard deviation of $X_i$; and $\sigma _{X_j}$ is the standard deviation of $X_j$.

The overall architecture of \ddcrnn is shown in Figure \ref{model-arch}. There are two inputs to the \ddcrnn model: 1) an adjacency matrix representing the traffic correlation among the sites of the WAN network topology and 2) the time-series data or the traffic statistic at each node of the graph. The encoder of the \ddcrnn network encodes the input into a fixed-length vector and passes it to the decoder.  The decoder forecasts future traffic conditions. During the training phase, the time series of $T'$ time steps on $N$ nodes is fed in as input; the correlation matrix is computed on the time-series data and given as the weighted adjacency matrix for diffusion convolution.  The encoder architecture takes the data and the decoder is used to forecast the output of the next $T$ time steps. The learnable weights of \ddcrnn are trained using a minibatch stochastic gradient, using a mean absolute error (MAE) loss function.

The \ddcrnn for R-WAN traffic forecasting is based on a DCRNN that was originally developed for forecasting traffic on highway networks \cite{li2017diffusion}.
The network traffic forecasting is similar to highway traffic forecasting, where the traffic diffuses from one node to the other node based on the network graph connectivity. However, the key difference stems from the spatial connectivity and temporal regularity. In a highway setting, the traffic exhibits regular patterns across weekdays, peak hours, regions (i.e., distribution of vehicles going from one region to another region), and seasons. However, such patterns cannot be seen in a R-WAN network because users can initiate large data transfers at any time. Therefore, \ddcrnn utilizes a dynamic adjacency matrix computed from the current state of the network traffic. During training, \ddcrnn learns to diffuse traffic on the graph under different weighted adjacency matrix settings. Consequently, during inference, the trained model can perform forecasts based on the time series of $T'$ steps and the weighted adjacency matrix computed on it.

\section{Experiments}

We used ESnet traffic traces for 1 year with 1-hour time resolution. We grouped data as follows: from 2018-01-01 to 2018-09-13 for training (70\%); from 2018-09-13 to 2018-10-20 (10\%) for validation; and from 2018-10-20 to 2018-12-31 (20\%) for testing. We measured 
the network traffic as bandwidth in GB/s. Each site has two bandwidth values: incoming and outgoing. Therefore, we mapped the 48 physical sites to a 96-node graph by considering two nodes (incoming and outgoing) for each physical site, as they have separate fiber optic links. Since data transfer varies significantly from site to site, we computed min-max scalar transformation on the training data and use it normalize bandwidth values of training, validation, and testing data. 




Our \ddcrnn implementation is based on the open-source DCRNN implementation \cite{li2017diffusion}. We used the following hyperparameter values for \ddcrnn: batch size: 64, number of epochs: 30, maximum diffusion steps: 2, number of RNN layers: 2, number of RNN units per layers: 16, max\_grad\_norm: 5, initial learning rate: 0.01, and learning rate decay: 0.1. These hyperparameter values were set as default for  the open-source DCRNN implementation.

The \ddcrnn training was performed on a single node of Cooley GPU cluster at Argonne Leadership Computing Facility. The node consists of two 2.4 GHz Intel Haswell E5-2620 v3 processors (6 cores per CPU, 12 cores total), one NVIDIA Tesla K80 (two GPUs per node), 384 GB of RAM per node, and 24 GB GPU RAM per node (12 GB per GPU). The software stack comprises Python 3.6.0, TensorFlow 1.3.1, NumPy 1.16.3, Pandas 0.19.2, and HDF5 1.8.17. The analysis was run on the Haswell compute nodes of Cori supercomputer at NERSC where each node has two sockets, and each socket is populated with a 2.3 GHz 16-core processor (Intel Xeon Processor E5-2698 v3) and 128GB DDR4 2133MHz memory.

To compare the forecasting accuracy of different models, we utilize mean absolute percentage error (MAPE) and coefficient of determination ($R^2$) computed on the original bandwidth scale (after applying inverse min-max scalar transformation).

\subsection{Exploratory analysis}

\begin{figure}[t]
  \centering
 \includegraphics[width=1\linewidth]{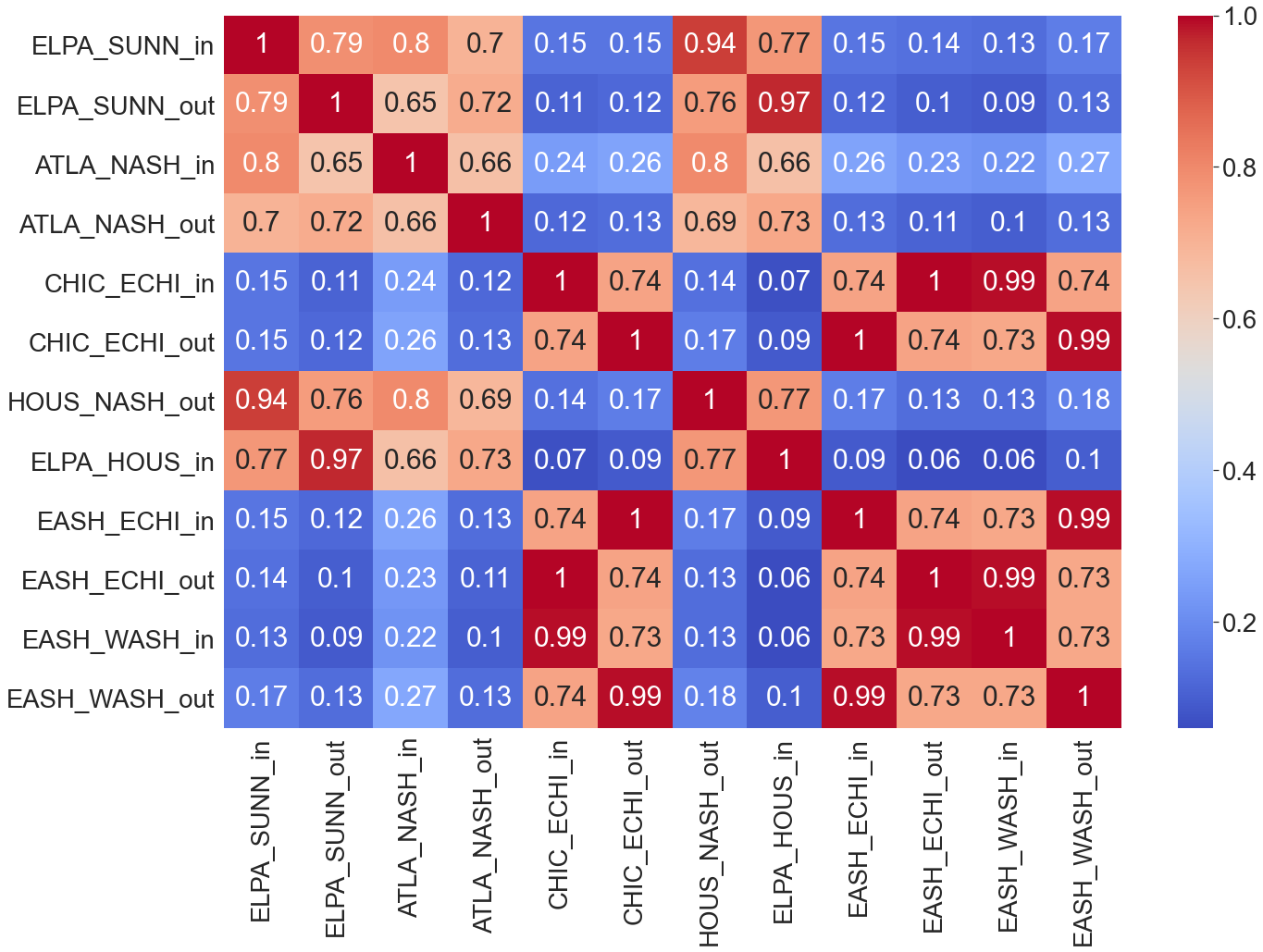}
  \caption{Pearson correlation coefficient matrix for the 12 nodes. These nodes exhibit high correlations and are selected for the exploratory analysis.} 
  \label{corryearplot}
\end{figure}


We conducted an exploratory analysis on \ddcrnn to study the impact of input horizon duration ($T'$), forecasting horizon ($T$), and the type of autoregression on the forecasting  accuracy. Given a pair of sites, we computed the Pearson correlation coefficient between their time series on the training data. We selected the 12 out of 96 sites based on the strong correlation values. The correlation matrix of the 12 sites are shown in Figure \ref{corryearplot}, where \_in and \_out refer to incoming and outgoing nodes, respectively. The reason for selecting the subset of nodes is to avoid bias in the experimental comparison. In particular, if we use all the nodes and select the best options, then it might introduce bias in favor of the \ddcrnn while comparing it with other methods. For the same reason, the forecasting accuracy for the exploratory analysis was computed on the validation data and not on the test data, thus avoiding test data leakage.



\subsubsection{Impact of input horizon duration}
\begin{table*}[t]
\centering
\begin{tabular}{|p{30mm}|l|l|l|l|l|l|l|l|}
\hline	
Input horizon duration & 6hrs & 12hrs & 18hrs & 24hrs & 30hrs & 36hrs & 42hrs & 48hrs   \\
\hline	
 \hline
$R^2$ ($\mu$) & 	0.58 & 	0.60 & 	0.70 & 0.72 & 	{\bf 0.76} & 	0.69	& 	0.71& 	0.77 \\ \hline			
$R^2$ ($\sigma$) &0.02 & 0.09 & 	0.10 & 	0.12	 & {\bf 0.07}	 & 0.11 & 0.05 & 0.08  \\ \hline

\hline	
 \hline
MAPE ($\mu$) & 	22.71 & 	22.60 & 	20.90 & 19.50 & 	{\bf 19.01} & 	22.38	& 	20.47& 	20.20 \\ \hline			
MAPE ($\sigma$) & 6.16 & 4.25 & 6.55 &1.74 & {\bf 4.13} & 8.23& 3.37 & 3.77 \\ \hline	\end{tabular}\caption{Mean ($\mu$) and standard deviation ($\sigma$) of $R^2$ and MAPE values for varying input horizon duration.}\label{r2mape24table}
\end{table*}



We studied the impact of the length of the input horizon ($T'$) on the forecasting accuracy of the \ddcrnn. For training \ddcrnn, we varied the length as 6, 12, 18, 24, 30, 36, 42, and 48 hours to forecast the traffic for the next 24 hours. 

The $R^2$ and MAPE values obtained on the validation data are shown in Table \ref{r2mape24table}. For each node, we considered all the observed and their corresponding predicted values from the \ddcrnn on the validation set and computed $R^2$ and MAPE. The mean and standard deviation values are computed over 12 nodes. We can observe that, up to 30 hours, the forecasting accuracy improves as the sequence length increases. However, after 30 hours, the improvements are not significant. Therefore, we use an input duration of 30 hours for the rest of the experiments. Note that a larger input horizons will eventually increase the training data size and time required for training.


\subsubsection{Impact of forecasting horizon}




We analyzed the impact of forecasting horizon on \ddcrnn performance.  Given 30 hours of input horizon, we trained the \ddcrnn model to forecast 6, 12, 18, 24 hours horizon. Note that by training the \ddcrnn model for 24 hours, one can get the forecasting results for all other horizons. However, this strategy may not be optimal for 6, 12, or 18 hours forecast. Therefore, we trained \ddcrnn for each forecasting horizon. 

Table \ref{forecast_mapetable} shows the $R^2$ and MAPE values for different forecasting horizons on the validation data. From the results, as expected, we observe a trend in which the increase in the forecasting horizon decreases the accuracy. We select 24 hours as a forecasting horizon for further study, where the \ddcrnn model will be trained to forecast 24 hours. However, we will analyze the forecasting accuracy for the intermediate time intervals as well.

\begin{table}[]
\centering
\begin{tabular}{|p{30mm}|c|c|c|c|}
\hline	
Forecasting Time Horizon & $R^2$ & MAPE \\ 
& $\mu$ ($\sigma$) & $\mu$ ($\sigma$)\\ \hline
\hline 
6hrs & 0.92 (0.07) & 11.31 (8.07)  \\ 			
12hrs &0.82 (0.10) & 16.74 (6.35)   \\ 
18hrs &0.72 (0.11) & 20.61 (6.55)   \\ 
24hrs &0.76 (0.07) & 19.01 (4.13)  \\ 
\hline	
\end{tabular}\caption{Mean ($\mu$) and standard deviation ($\sigma$) of $R^2$ and MAPE values for varying  forecasting horizon.}
\label{forecast_mapetable}
\end{table}

\subsubsection{Comparison between autoregressive and nonautoregressive \ddcrnn}
We compared the forecasting accuracy of autoregressive and nonautoregressive \ddcrnn variants. By default, \ddcrnn adopts a nonautoregressive approach to forecasting. An alternative approach is autoregressive forecasting, where \ddcrnn can be trained to forecast only a one-time step, which is then given to the model recursively to obtain forecasting for 24 hours. This study is motivated by a previous work \cite{10.1145/3391812.3396268}, where autoregressive forecasting was adopted within LSTM models for traffic forecasting.


\begin{figure}
  \centering
    \includegraphics[width=1.0\linewidth]{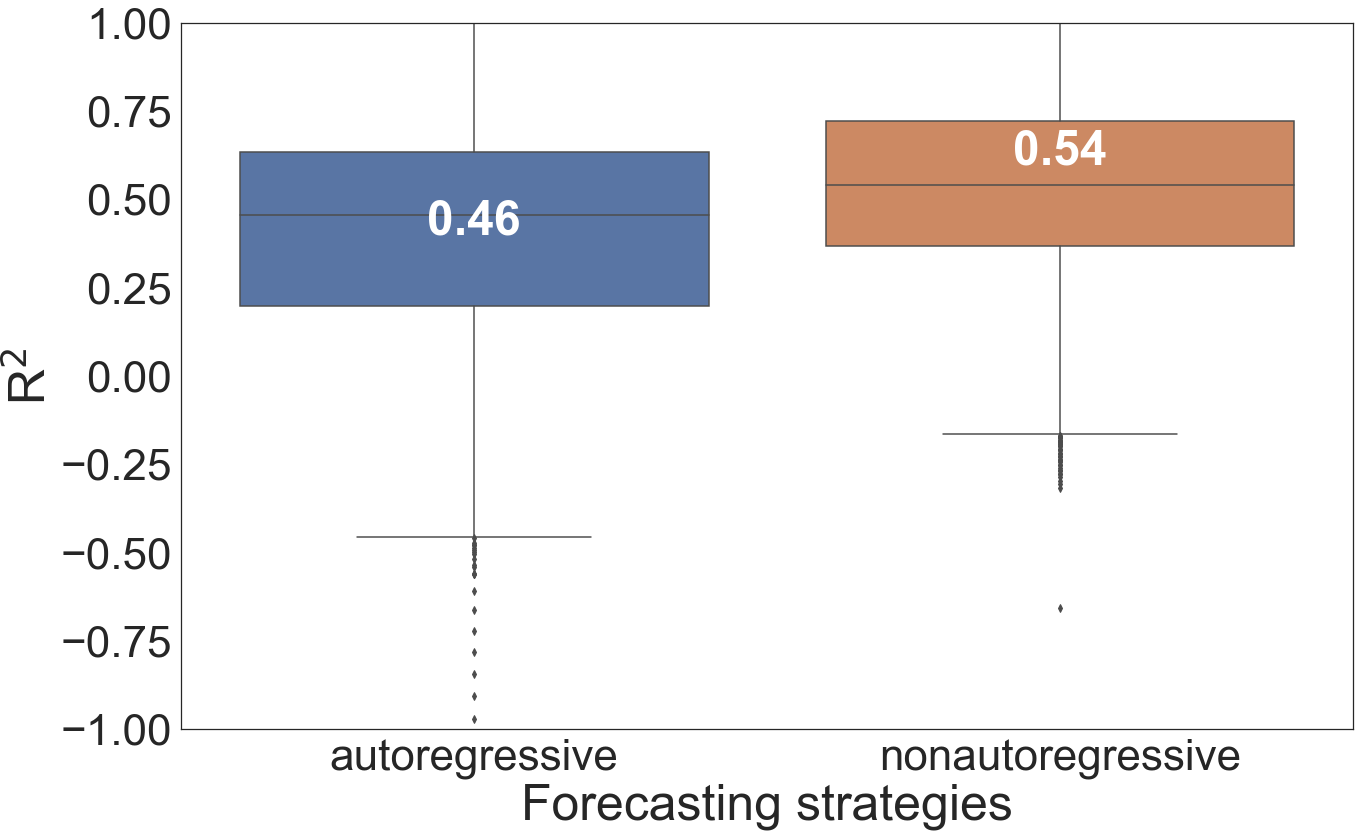}
  \caption{Distribution of R$^2$ values obtained by autoregressive and nonautoregressive \ddcrnn variants for the 24$^{th}$ hour forecast.}
  \label{fig_multi_step1}
\end{figure}

\begin{figure}
  \centering
    \includegraphics[width=1.0\linewidth]{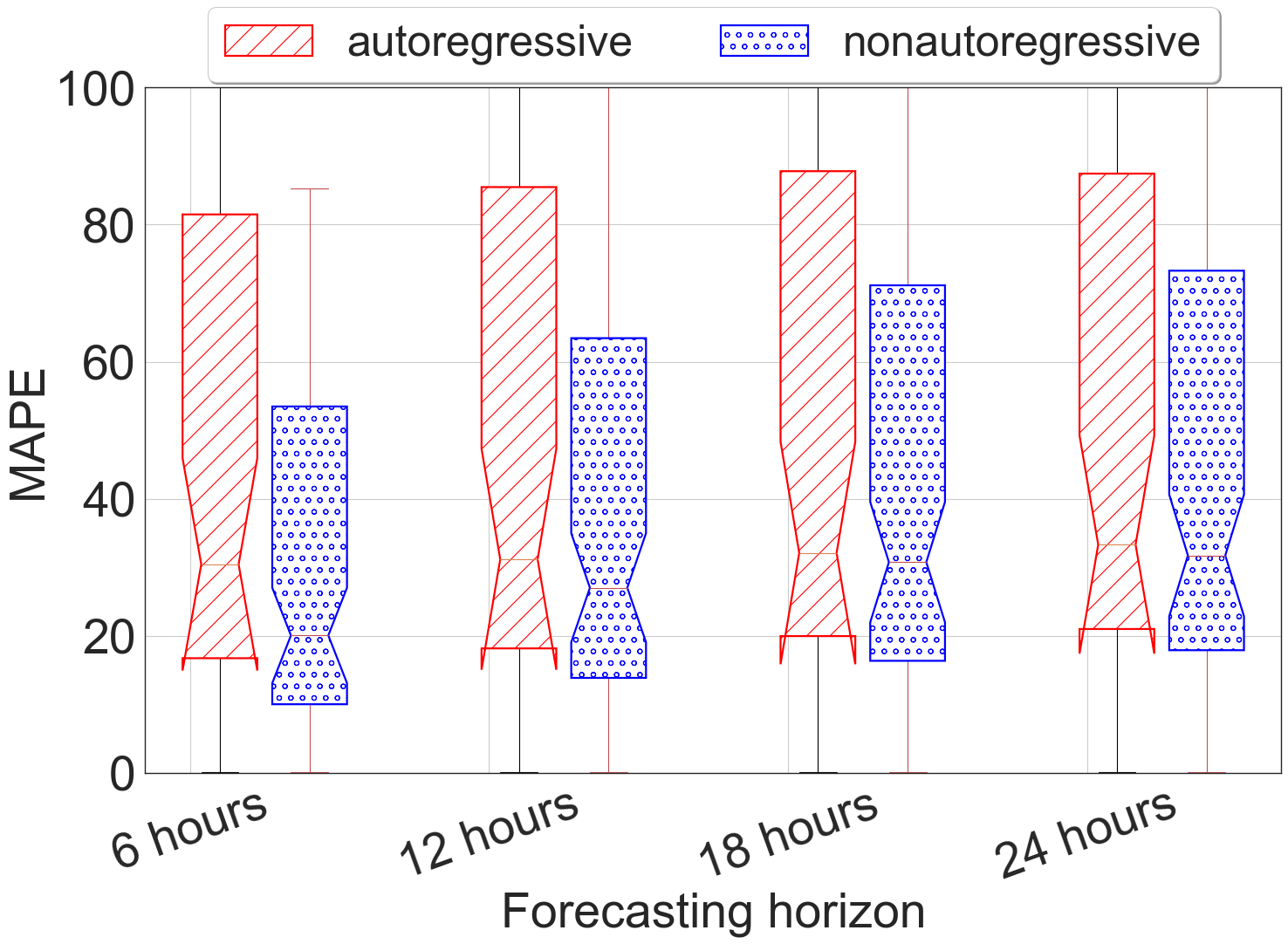}
  \caption{Distribution of MAPE values obtained by autoregressive and nonautoregressive \ddcrnn variants for different forecasting intervals.}
  \label{fig_multi_step}
\end{figure}

\begin{figure}[t]
     \centering
     \begin{subfigure}[b]{0.5\textwidth}
         \centering
         \includegraphics[width=\textwidth]{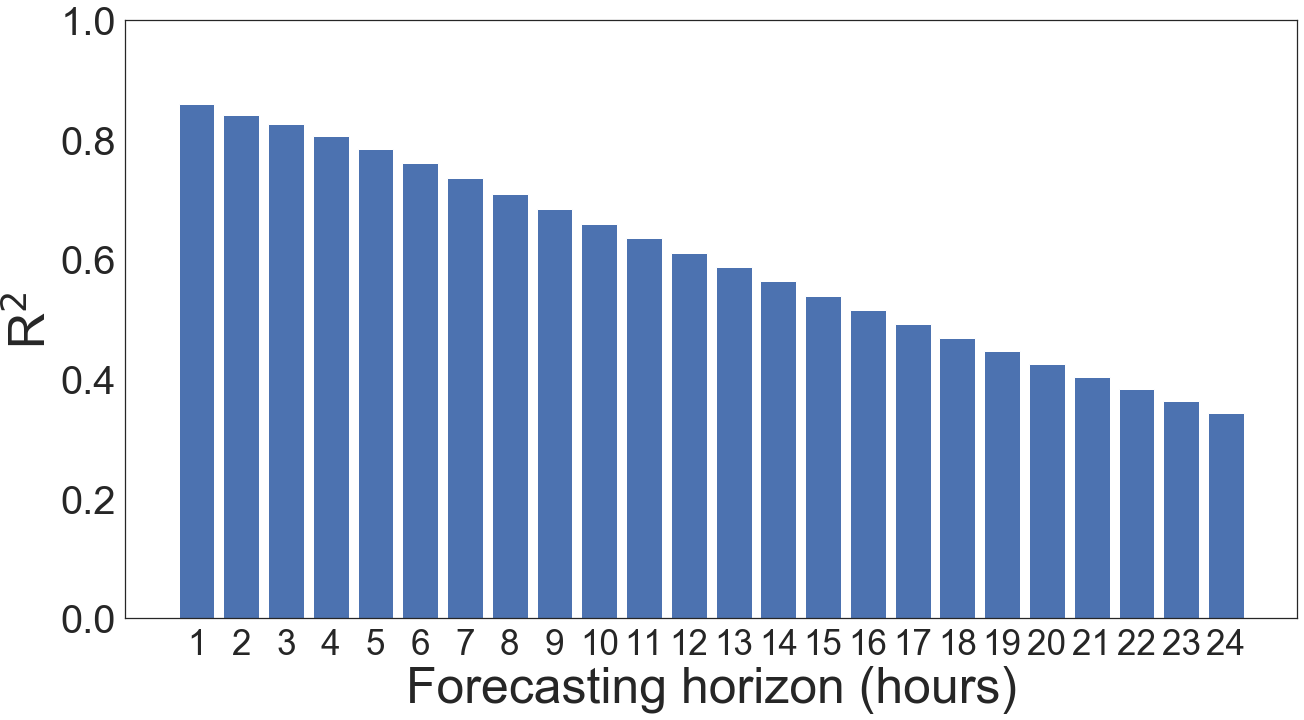}
         \caption{Autoregressive forecasting strategy}
         \label{fig_recursive}
     \end{subfigure}
     \hfill
     \begin{subfigure}[b]{0.5\textwidth}
         \centering
         \includegraphics[width=\textwidth]{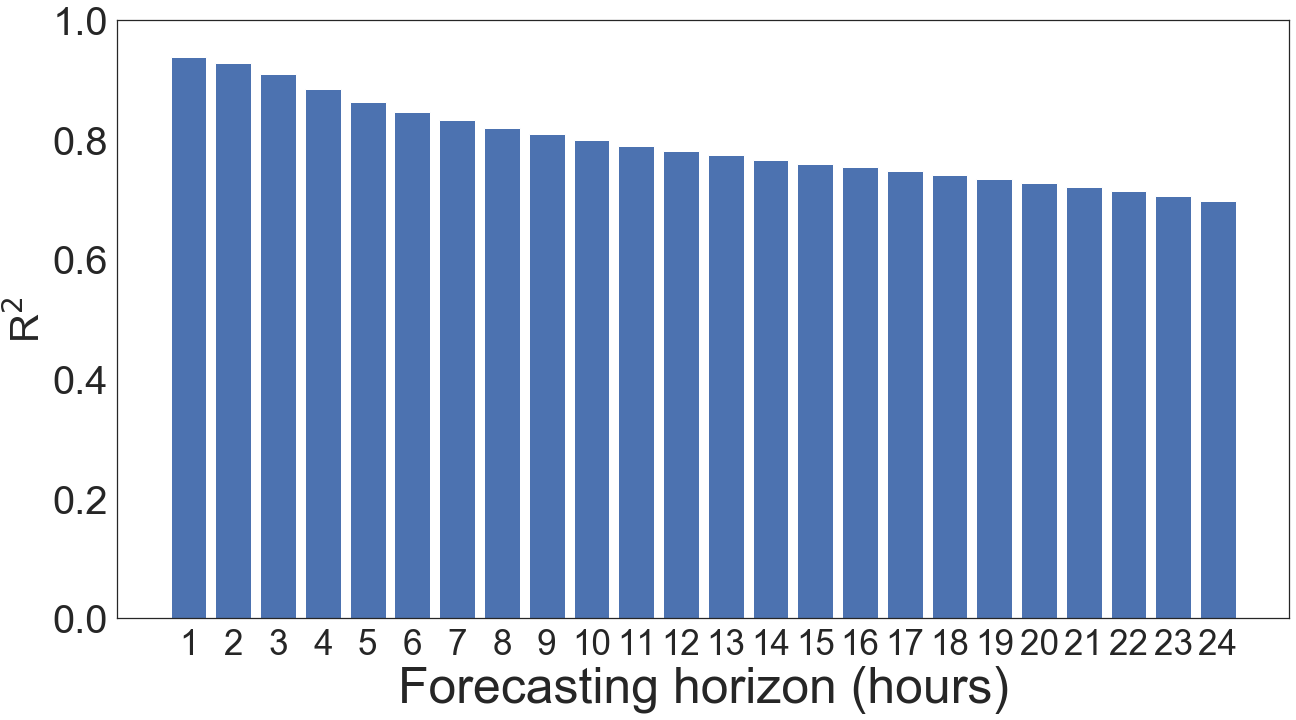}
         \caption{Nonautoregressive forecasting strategy}
         \label{fig_multi_output}
    \end{subfigure}
    \caption{Comparison between autoregressive and nonautoregressive \ddcrnn variants on  ELPA\_SUNN\_in site.}
    \label{fig_forecast_strat}
\end{figure}


Figure \ref{fig_multi_step1} shows the distribution of $R^2$ values obtained by the two strategies at the 24$^{th}$ hour forecast. Figure \ref{fig_multi_step} shows the distribution of MAPE values for different forecasting horizon intervals. The result shows that the default nonautoregressive strategy performs significantly better than the autoregressive strategy. To take a close look, we plot the forecasting accuracy ($R^2$) of a particular node ELPA\_SUNN\_in Figure \ref{fig_forecast_strat}. The results show that the forecasting accuracy of autoregressive \ddcrnn decreases rapidly with an increase in forecasting time steps. This can be attributed to the error introduced at each time step, which drastically reduces accuracy for the next time step. This issue is less severe in the nonautoregressive strategy because it was trained on the entire forecasting horizon.


\subsection{Comparison between \ddcrnn and DCRNN}
Here, we compare the forecasting accuracy of  \ddcrnn and DCRNN to show that our dynamic approach that we proposed is critical for forecasting accuracy. These two methods differ only with respect to how the weighted adjacency matrix is used in the diffusion convolution. While in the former it is dynamically computed based on the current traffic state, in the latter it is static and based on the connectivity and distance. We also note that DCRNN has been applied to network traffic forecasting previously in \cite{8845132}. Both \ddcrnn and DCRNN used the same hyperparameters and the same training data with 96 nodes. We used 30 hours of input horizon to forecast 24 hours of output horizon. The forecasting accuracy values are computed on the test data.

\begin{figure}
  \centering
    \includegraphics[width=1.0\linewidth]{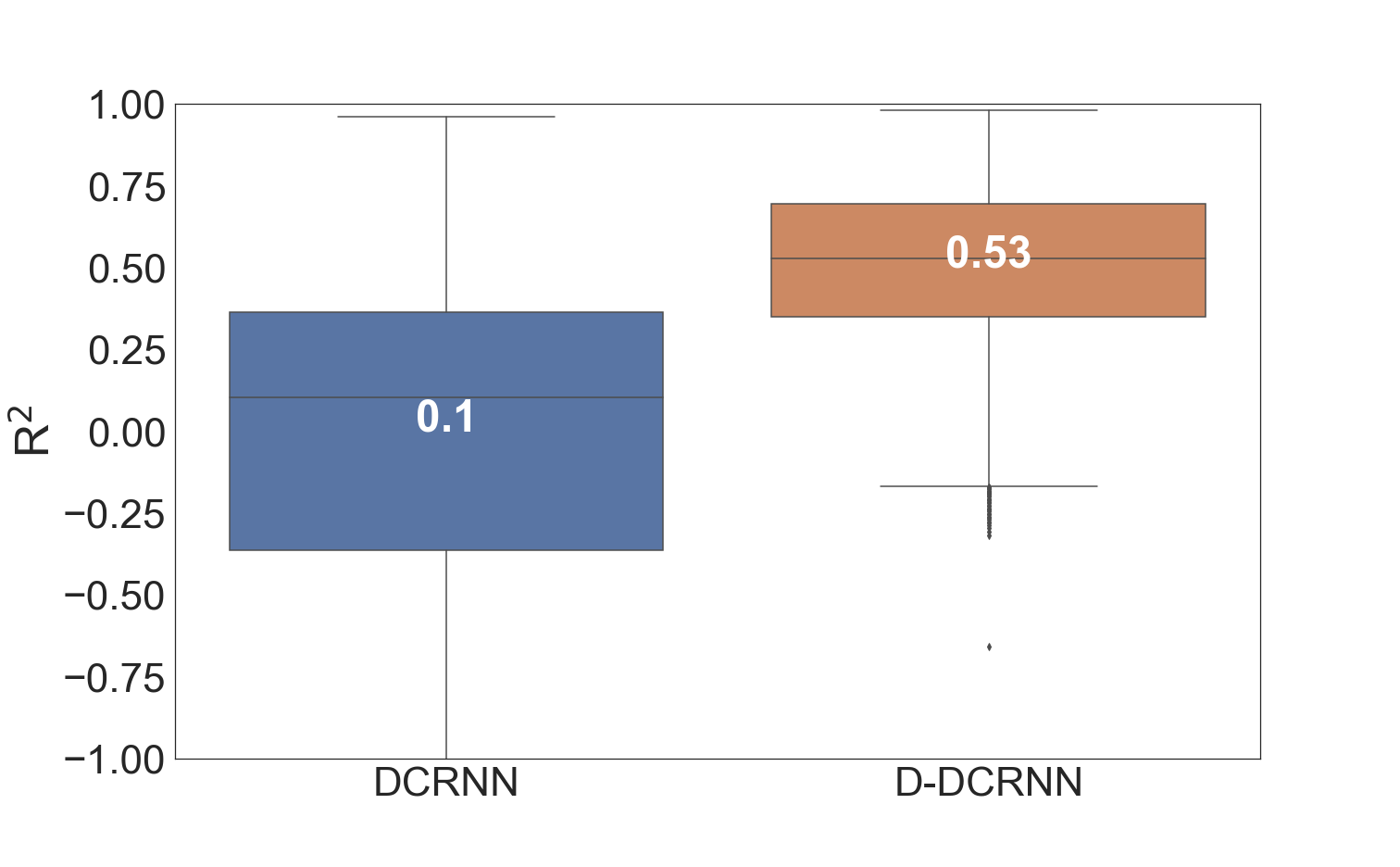}
  \caption{Distribution of R$^2$ values obtained by \ddcrnn and DCRNN for the 24$^{th}$ hour forecast.}
  \label{fig_dcrnn1}
\end{figure}

\begin{figure}
  \centering
    \includegraphics[width=1.0\linewidth]{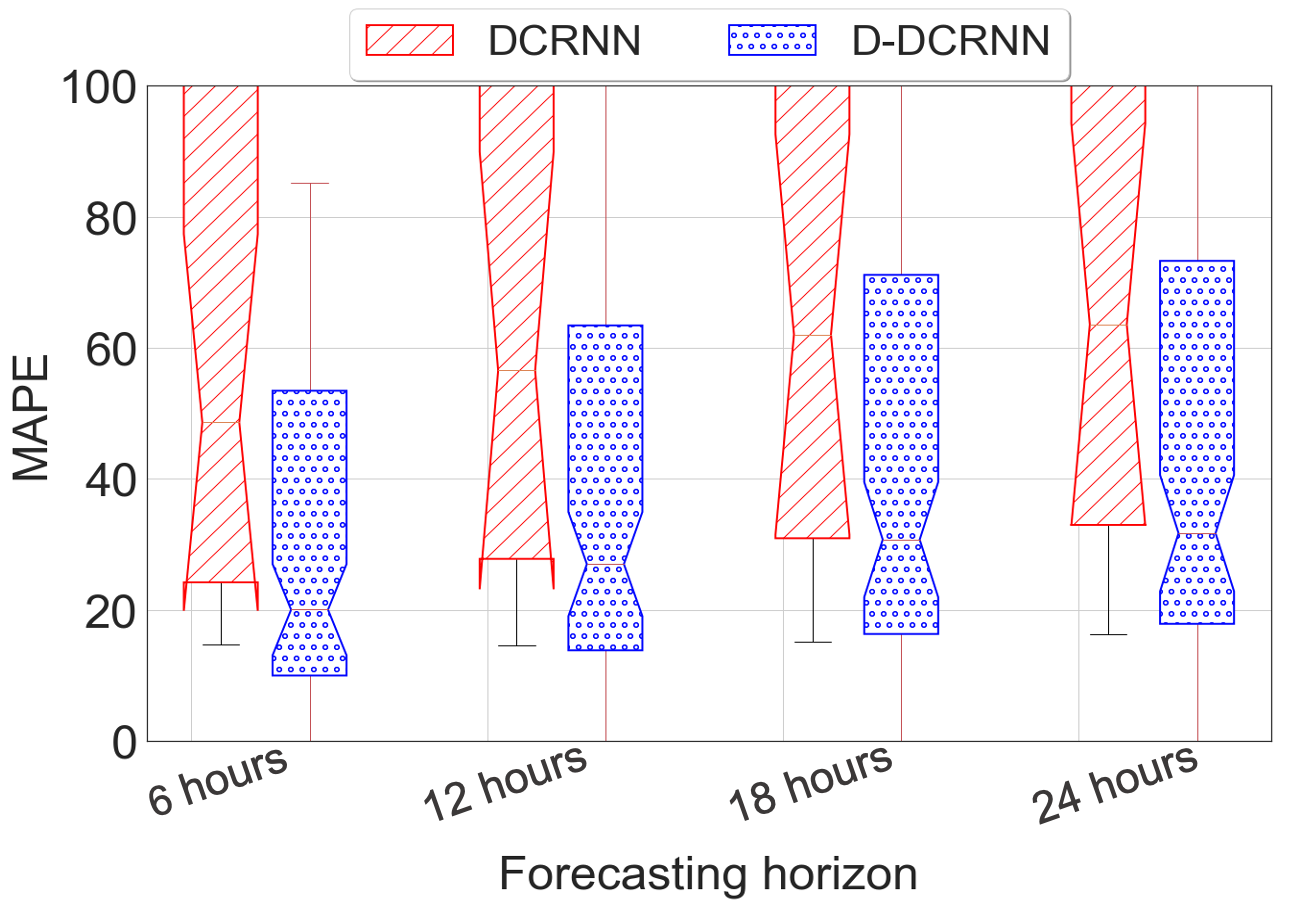}
  \caption{Distribution of MAPE values obtained by \ddcrnn and DCRNN for different forecasting intervals.}
  \label{fig_dcrnn2}
\end{figure}

Figure \ref{fig_dcrnn1} shows the $R^2$ distribution obtained by \ddcrnn and DCRNN for the 24$^{th}$ hour forecast. Figure \ref{fig_dcrnn2} shows the MAPE distribution for different forecasting horizon intervals. We can observe that \ddcrnn achieves forecasting accuracy, which is significantly better than DCRNN for all forecasting horizons. The superior performance of \ddcrnn can be attributed to its ability to model the dynamic spatiotemporal traffic patterns; DCRNN did not have this capability because it uses the static adjacency matrix, which leads to poor forecasting accuracy. While a previous work \cite{8845132} reported superior performance of the direct application of DCRNN to network traffic, we hypothesize that it was due to less dynamic traffic data with shorter forecasting horizon.


\subsection{Comparison of DDCRNN with other methods}
Here, we compare \ddcrnn with other statistical, machine learning, and deep learning methods including those previously studied for network traffic forecasting, and show that \ddcrnn outperforms all these methods. 

We used ARIMA \cite{williams2003modeling} (with order (2,1,2) implemented using statsmodel \cite{seabold2010statsmodels} Python package), linear regression (LR) \cite{lau2008local} (from scikit-learn), Random Forest (RF) \cite{dudek2015short} (from scikit-learn with default hyperparameters), Gradient Boosting (GB) \cite{taieb2014gradient} (from scikit-learn with default hyperparameters), Simple RNN (SRNN) \cite{laptev2017time} with one hidden layers with 4 neurons (implemented in Keras), LSTM \cite{selvin2017stock}, stacked LSTM (SLSTM): two hidden layers with 256 neurons per layer (implemented in Keras), Gated Recurrent Unit \cite{fu2016using}: one hidden layers with 4 neurons (implemented in Keras), and FireTS \cite{xie2018benchmark,xie2020}, a Python package for multivariate time series forecasting with linear regression as base estimator, auto-regression order of 24, exogenous order of 24, and exogenous delay of 0. For each node, we build a node-specific model using each method. Consequently, for a given method, we have 96 models. For \ddcrnn, we have a single model to forecast the traffic on all 96 nodes. We used 30 hours of input horizon to forecast 24 hours of output horizon. We computed the accuracy values on the test data.

\begin{figure*}[t]
     \centering
     \begin{subfigure}[b]{0.32\textwidth}
         \centering
         \includegraphics[width=\textwidth]{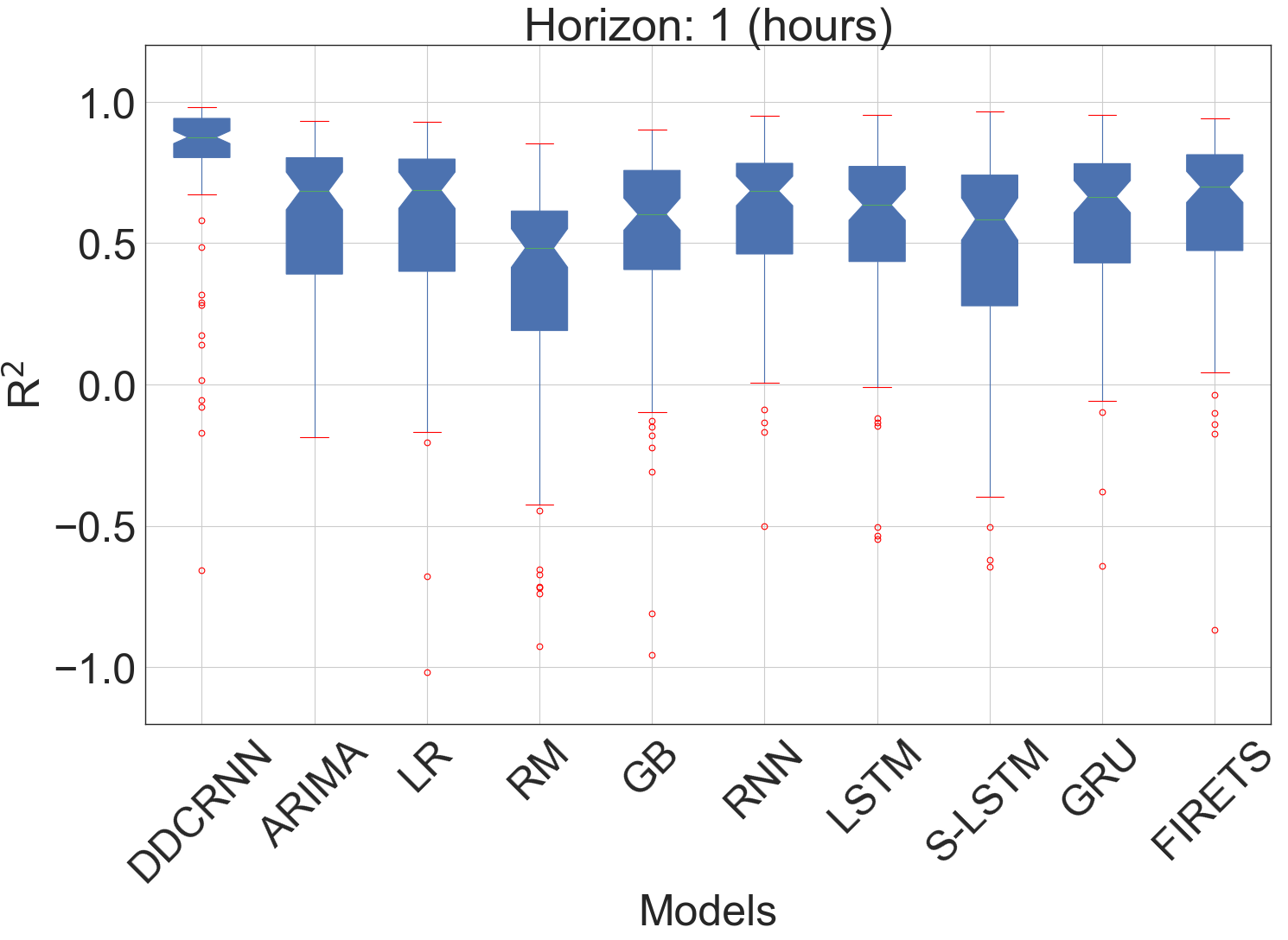}
         \caption{$1^{st}$ hour forecasting}
         \label{fig_comp_1hr}
     \end{subfigure}
     \hfill
     \begin{subfigure}[b]{0.32\textwidth}
         \centering
         \includegraphics[width=\textwidth]{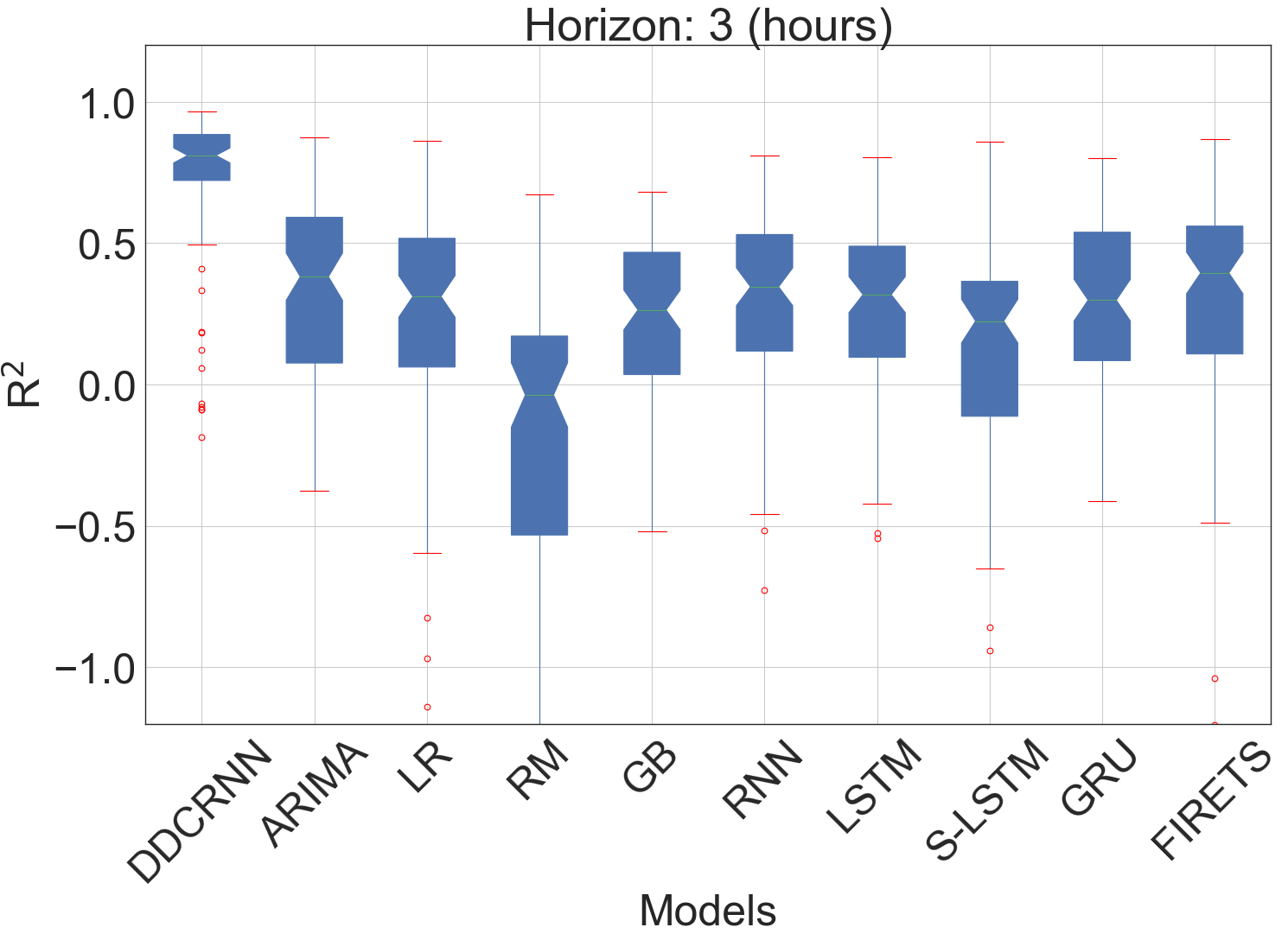}
         \caption{$3^{rd}$ hour forecasting}
         \label{fig_comp_3hr}
     \end{subfigure}
     \hfill
     \begin{subfigure}[b]{0.32\textwidth}
         \centering
         \includegraphics[width=\textwidth]{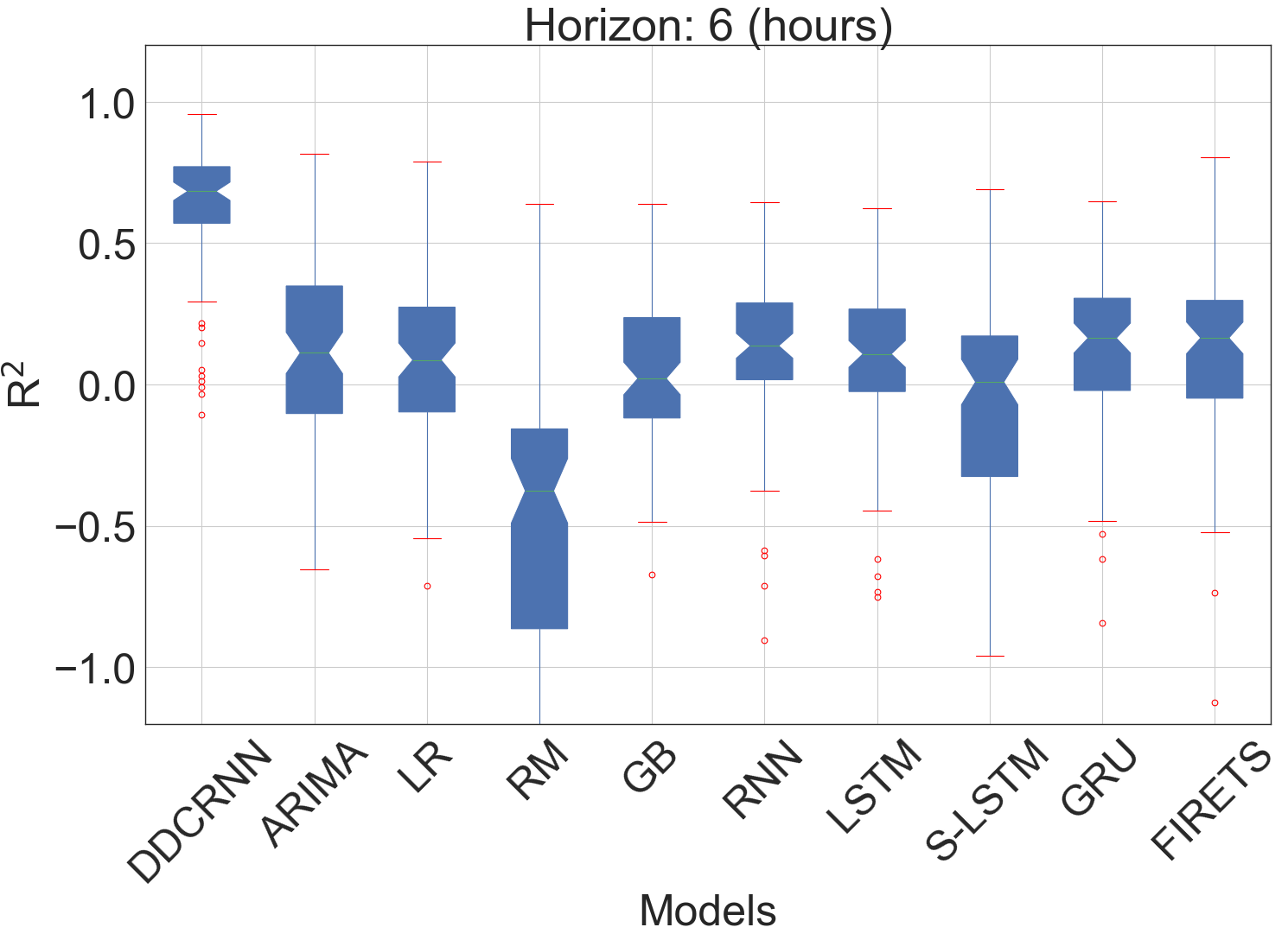}
         \caption{$6^{th}$ hour forecasting}
         \label{fig_comp_6hr}
     \end{subfigure}
     \hfill
     \begin{subfigure}[b]{0.32\textwidth}
         \centering
         \includegraphics[width=\textwidth]{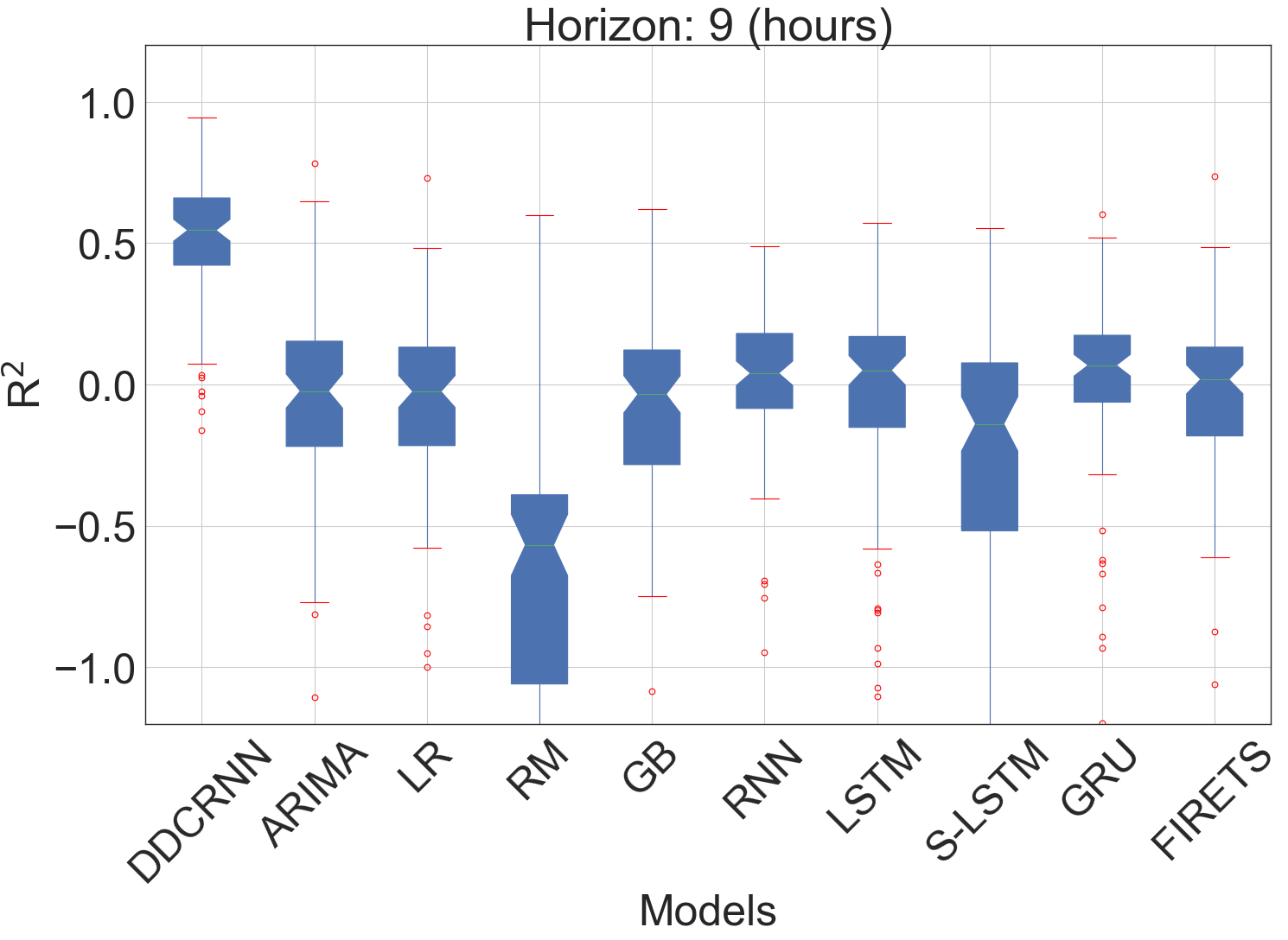}
         \caption{$9^{th}$ hour forecasting}
         \label{fig_comp_9hr}
     \end{subfigure}
     \hfill
     \begin{subfigure}[b]{0.32\textwidth}
         \centering
         \includegraphics[width=\textwidth]{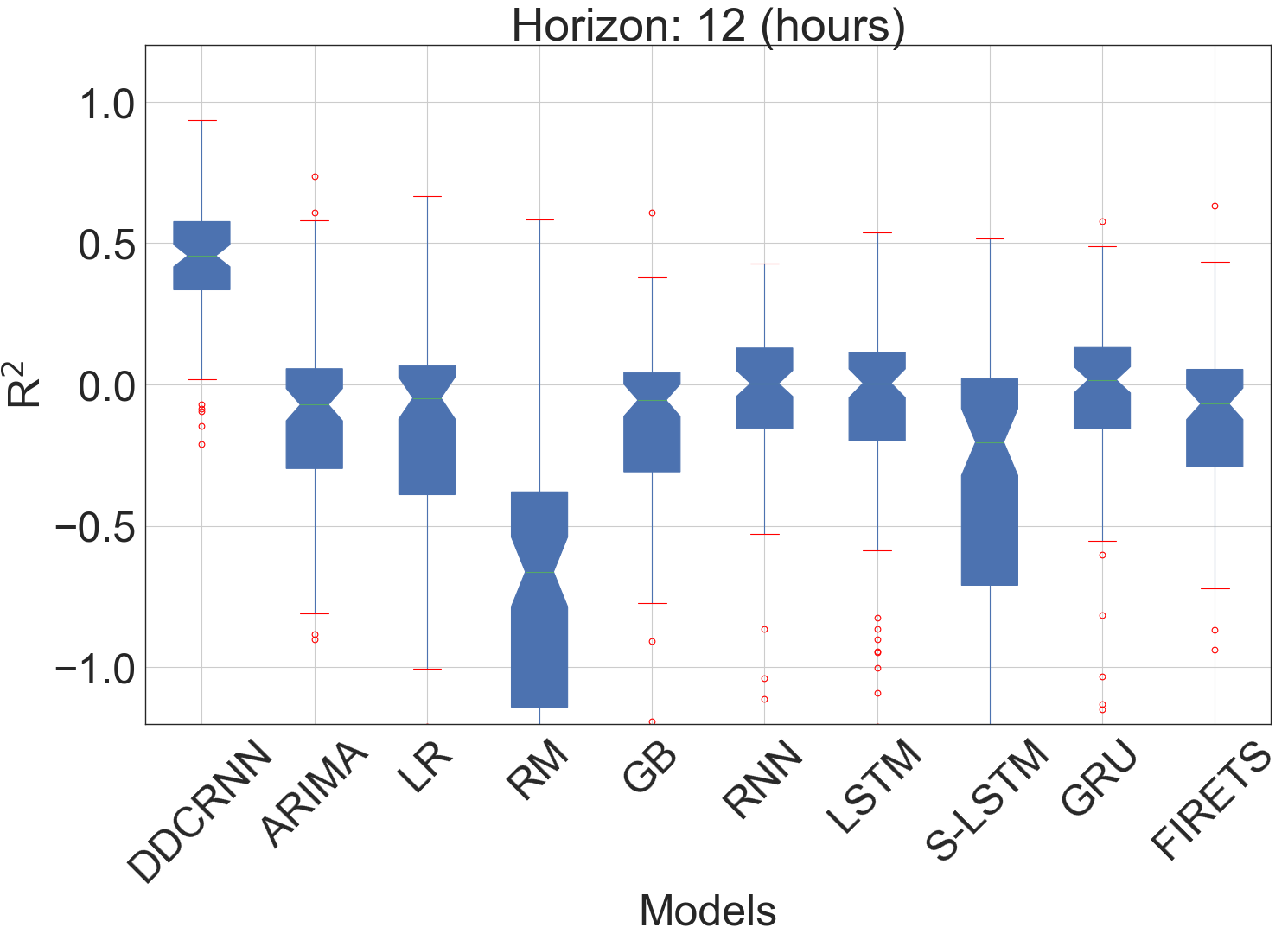}
         \caption{$12^{th}$ hours forecasting}
         \label{fig_comp_12hr0}
     \end{subfigure}
     \hfill
     \begin{subfigure}[b]{0.32\textwidth}
         \centering
         \includegraphics[width=\textwidth]{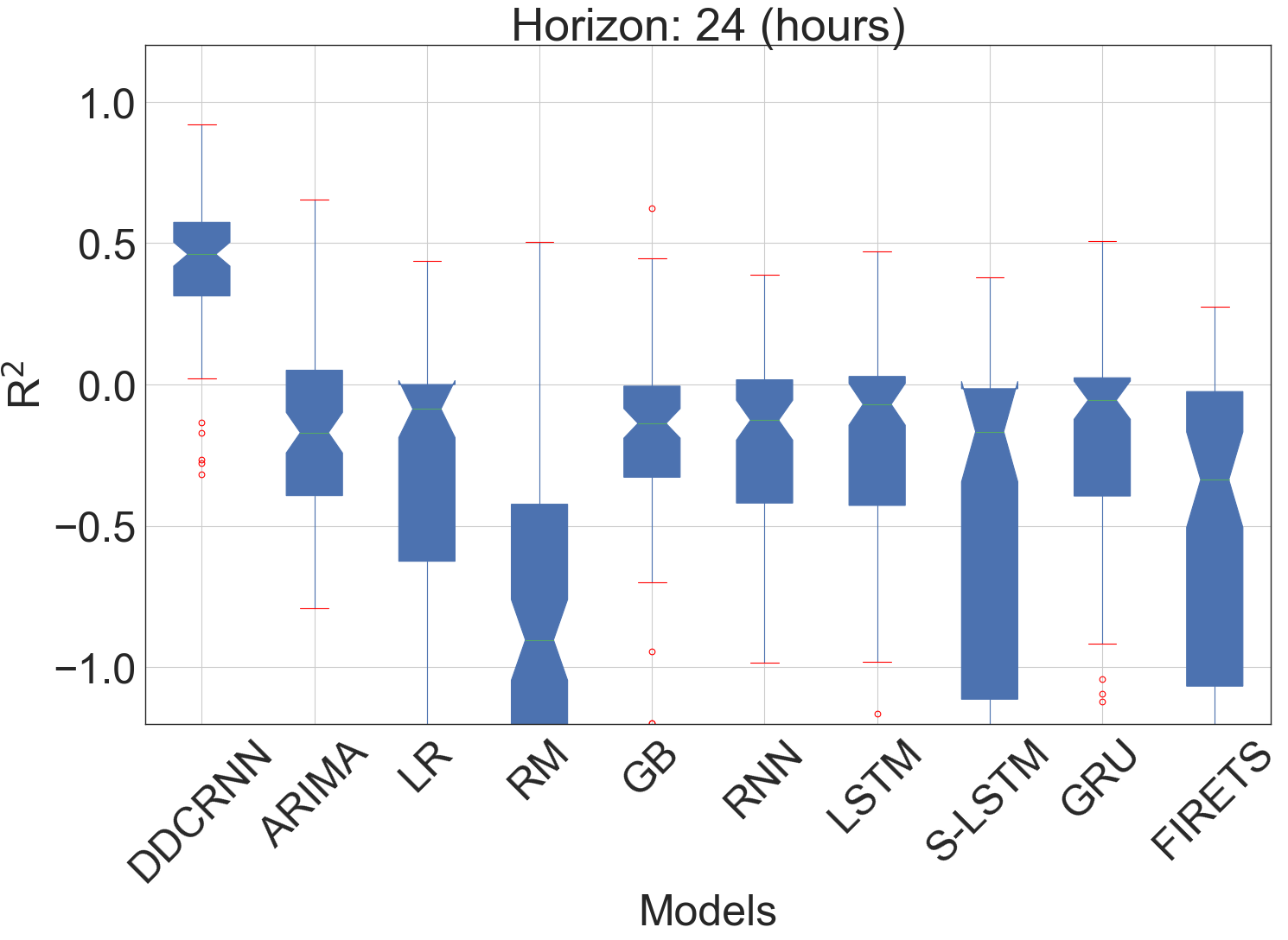}
         \caption{$24^{th}$ hour forecasting}
         \label{fig_24hr_forcast}
     \end{subfigure}
        \caption{Distribution of $R^2$ values obtained on 96 nodes by \ddcrnn and other node-specific models for different forecasting horizon intervals.}
        \label{fig_comp_different_hours}
\end{figure*}

Figure \ref{fig_comp_different_hours} shows the distribution of $R^2$ obtained by different methods on the test data for different forecasting intervals (1, 3, 6, 9, 12, and 24 hours). The results show that \ddcrnn achieves forecasting accuracy values that are significantly higher than all the other node-specific models. At 1$^{st}$ hour forecast, node-specific models accuracy values are closer to \ddcrnn, however, they become poor with an increase in the forecasting horizon. Starting from the 3$^{rd}$ hour forecast, \ddcrnn results are significantly better than the node-specific models. From 6$^{th}$ hour forecast, we can observe that only \ddcrnn has achieved reasonable accuracy. The superior performance of \ddcrnn over node-specific models can be attributed to the former's  ability to capture both spatial and temporal patterns using gated recurrent units with diffusion convolution defined on a dynamic graph.


\begin{figure*}[t]
     \centering
          \begin{subfigure}[b]{0.30\textwidth}
         \centering
         \includegraphics[width=\textwidth]{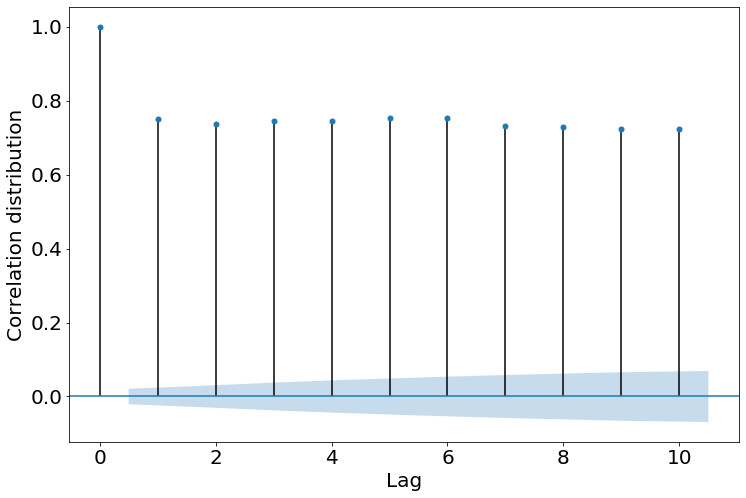}
         \caption{ACF for ANL-STAR-in}
         \label{fig_comp_1hr-acf}
     \end{subfigure}
     \hfill
     \begin{subfigure}[b]{0.30\textwidth}
         \centering
         \includegraphics[width=\textwidth]{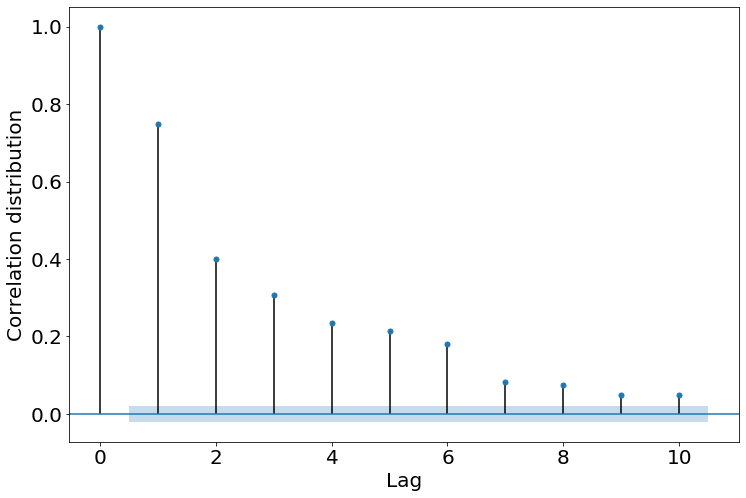}
         \caption{PACF for ANL-STAR-in}
         \label{fig_comp_3hr-pacf}
     \end{subfigure}
     \hfill
     \begin{subfigure}[b]{0.30\textwidth}
         \centering
         \includegraphics[width=\textwidth]{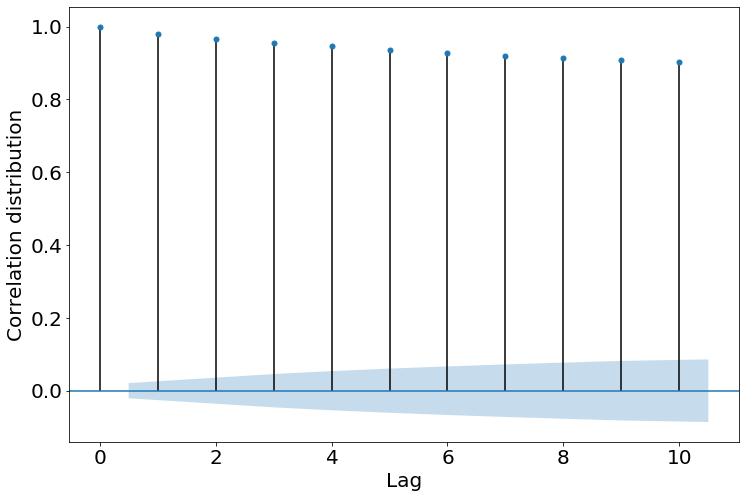}
         \caption{ACF for SLAC-SUNN-in}
     \end{subfigure}
     \hfill
     \begin{subfigure}[b]{0.30\textwidth}
         \centering
         \includegraphics[width=\textwidth]{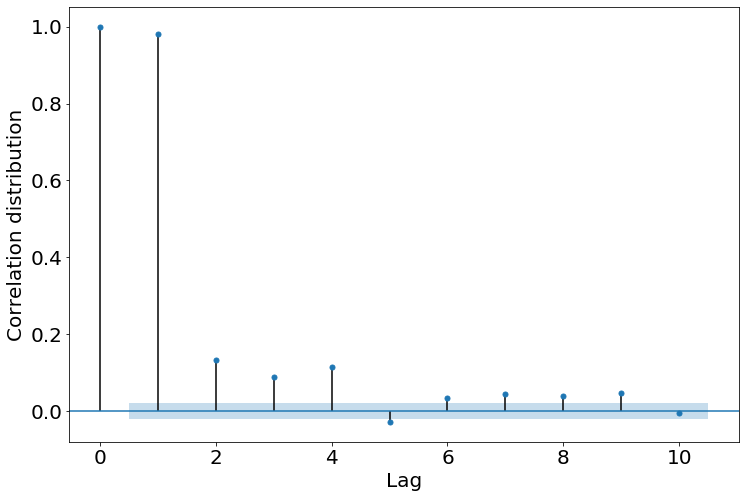}
         \caption{PACF for SLAC-SUNN-in}
         \label{fig_pacf_slac}
     \end{subfigure}
    \hfill
     \begin{subfigure}[b]{0.30\textwidth}
         \centering
         \includegraphics[width=\textwidth]{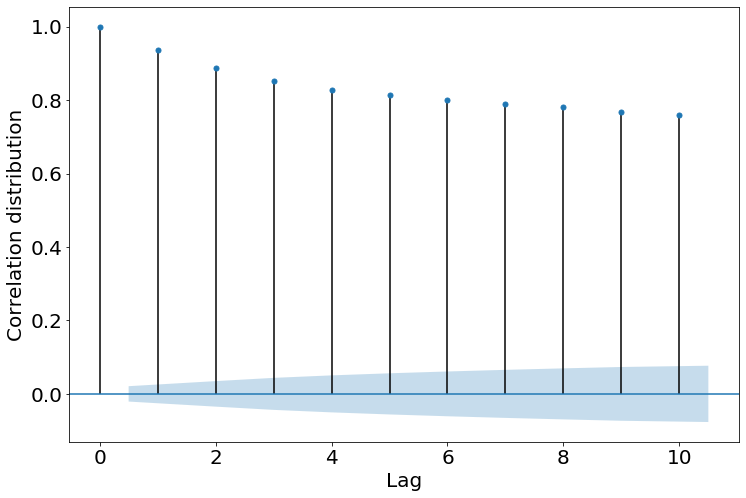}
         \caption{ACF for JGI-SACR-in}
         \label{fig_acf_jgi}
     \end{subfigure}
     \hfill
     \begin{subfigure}[b]{0.30\textwidth}
         \centering
         \includegraphics[width=\textwidth]{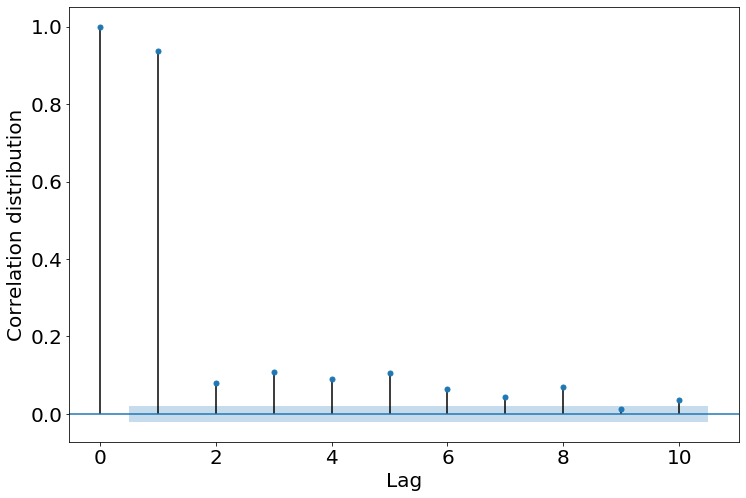}
         \caption{PACF for JGI-SACR-in}
      \label{fig_pacf_jgi}
     \end{subfigure}
     \hfill
     \begin{subfigure}[b]{0.30\textwidth}
         \centering
         \includegraphics[width=\textwidth]{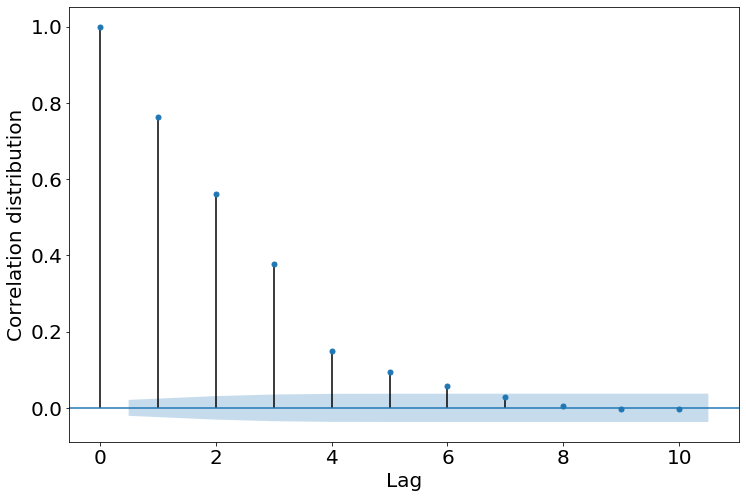}
         \caption{ACF for BNL-NEWY-in}
         \label{acfanlstarin}
     \end{subfigure}
     \hfill
     \begin{subfigure}[b]{0.30\textwidth}
         \centering
         \includegraphics[width=\textwidth]{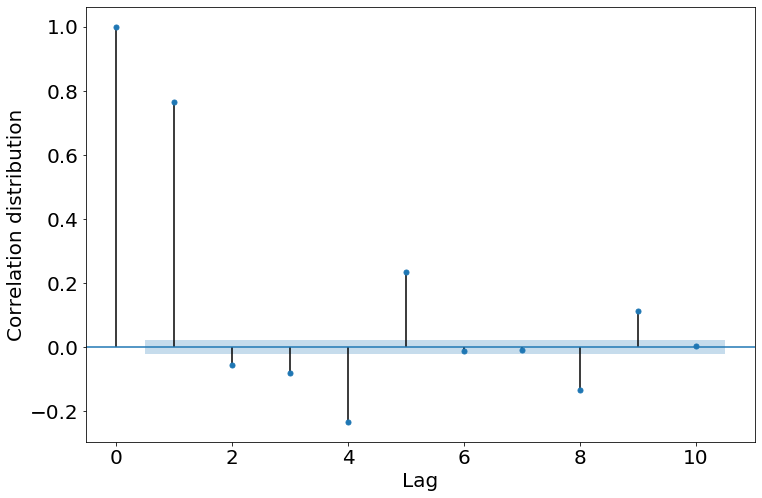}
         \caption{PACF for BNL-NEWY-in}
         \label{pacfanlstain}
     \end{subfigure}
     \hfill
     \begin{subfigure}[b]{0.30\textwidth}
         \centering
         \includegraphics[width=\textwidth]{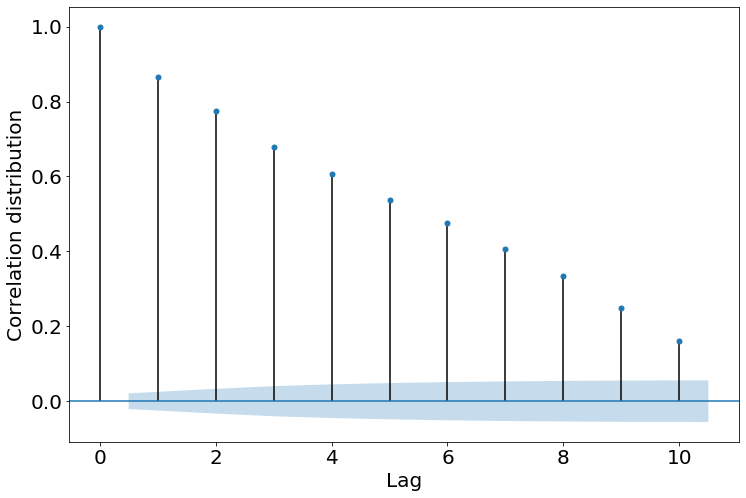}
         \caption{ACF for ALBQ-DENV-out}
     \end{subfigure}
     \hfill
     \begin{subfigure}[b]{0.30\textwidth}
         \centering
         \includegraphics[width=\textwidth]{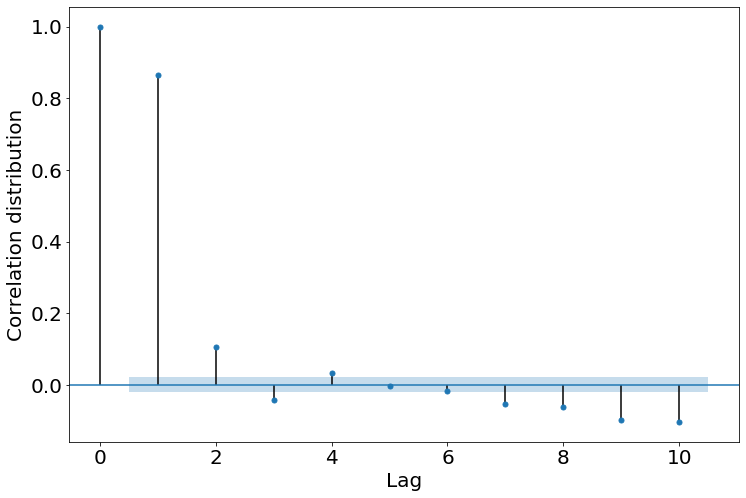}
         \caption{PACF for ALBQ-DENV-out}
         \label{fig_pacf_albq}
     \end{subfigure}
          \hfill
     \begin{subfigure}[b]{0.30\textwidth}
         \centering
         \includegraphics[width=\textwidth]{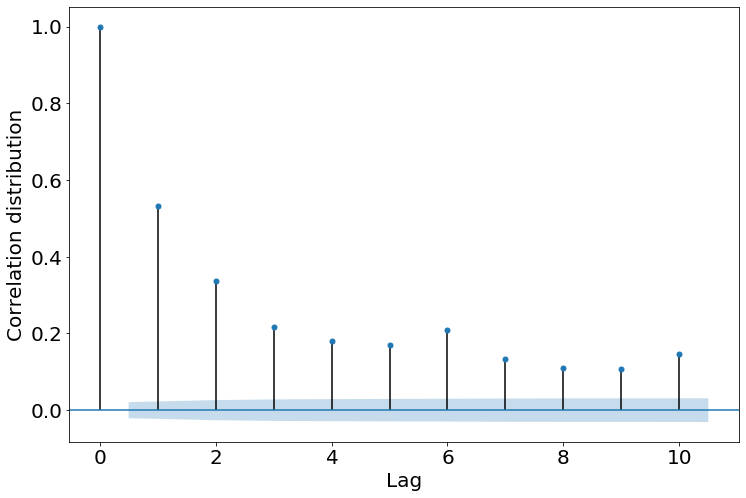}
         \caption{ACF for CERN513-WASH-in}
         \label{fig_acf_cern}
     \end{subfigure}
     \hfill
     \begin{subfigure}[b]{0.30\textwidth}
         \centering
         \includegraphics[width=\textwidth]{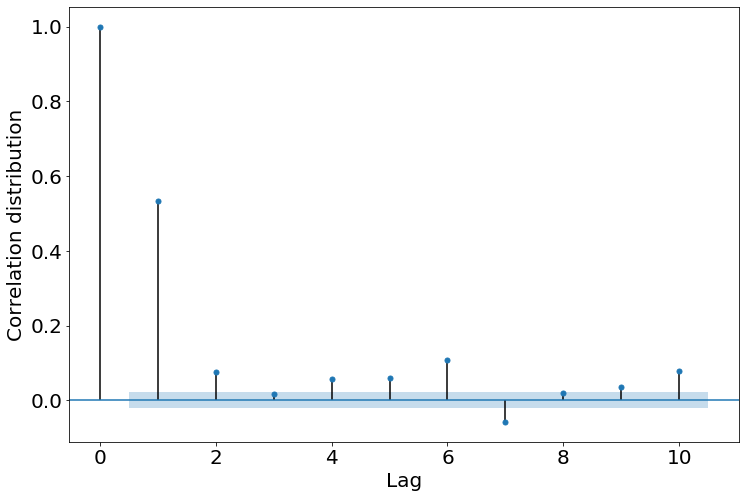}
         \caption{PACF for CERN513-WASH-in}
      \label{fig_pacf_cern}
     \end{subfigure}
        \caption{Autocorrelation (ACF) and partial autocorrelation (PACF) plot showing correlation of observations as a function of time lag for different sites.}
        \label{fig_good_bad}
\label{autocorr}
\end{figure*}

\subsection{Analysis and characterization}

Here, we analyse the forecasting accuracy of \ddcrnn and provide insights on its strengths and limitations.

Table \ref{goodbad} shows 10 nodes where \ddcrnn obtained best and worst $R^2$ values, respectively. For each node, we computed the autocorrelation and partial autocorrealtion values with a lag of 10 to analyse to what extent the values in the time series are correlated. Then, we computed the mean of the obtained autocorrelation values. Table \ref{forecast_mapetable} shows the mean autocorrelation values for the top three high and low accuracy nodes. For the same nodes, the autocorrelation and partial autocorrelation plots are show in Figure \ref{autocorr}.

\begin{table}[!ht]
\centering
\begin{tabular}{|c|c|c|}
\hline
S/n     & High accuracy & Low accuracy   \\ \hline
1 & ANL\_STAR\_in &  BNL\_NEWY\_in \\ \hline
2 & ANL\_STAR\_out &  ALBQ\_DENV\_out \\ \hline
3 & SLAC\_SUNN\_in &  CERN-513\_WASH\_in \\ \hline
4 & JGI\_SACR\_in &  NASH\_WASH\_in\\ \hline
5 & FNAL\_STAR\_out &  CERN-513\_WASH\_out \\ \hline
6 & BNL\_NEWY\_out &  ATLA\_ORNL\_out \\ \hline
7 & NERSC\_SUNN\_out &  NASH\_WASH\_out\\ \hline
8 & NERSC\_SUNN\_in &   ATLA\_SRS\_out \\ \hline
9 & SLAC\_SUNN\_out &  ATLA\_NASH\_in \\ \hline
10 & LSVN\_SUNN\_out &  ELPA\_HOUS\_out\\ \hline
\end{tabular}\caption{Nodes with high and low forecasting accuracy.}\label{goodbad}
\end{table}

\begin{table}[]
\centering
\begin{tabular}{|p{30mm}|c|c|c|}
\hline	
Sites & Mean ACF Value  \\ \hline
\hline 
\multicolumn{2}{|c|}{High accuracy}\\
\hline
ANL\_STAR\_in  & 0.783  \\ 			
SLAC\_SUNN\_in & 0.974   \\ 
JGI\_SACR\_in  & 0.842    \\ 
\hline
\multicolumn{2}{|c|}{Low accuracy}\\
\hline
BNL\_NEWY\_in  & 0.238   \\ 
ALBQ\_DENV\_out& 0.526  \\ 
CERN-513\_WASH\_in & 0.232  \\ 
\hline	
\end{tabular}\caption{Mean autocorrelation values for the top three nodes with high and low accuracy.}
\label{forecast_mapetable}
\end{table}

From the results, we observed that nodes, where \ddcrnn obtained high accuracy, exhibit large autocorrelation values. On the other hand, \ddcrnn obtains poor results when there is no strong temporal correlation, where the traffic is random. Upon further analysis, we found that nodes with high forecasting accuracy such as ANL\_STAR\_in, ANL\_STAR\_out, and SLAC\_SUNN\_in 
have large experimental facilities. These include Argonne National Laboratory, Stanford Linear Accelerator, Joint Genome Institute, and Linac Coherent Light Source.
In these user facilities, the data transfers can happen any time, however, when a transfer happens, a large volume of experimental data will be moved to data centers for further analysis. These transfers can last several hours and thus exhibit spatial temporal patterns. \ddcrnn can learn these patterns to provide high forecasting accuracy.   
Specifically, our analysis reveal that \ddcrnn can model short-term transfers, particularly long connections leading to predictable bulk data transfers especially driven by these large facilities. With respect to low accuracy nodes, we see a multitude of factors that prevented \ddcrnn from reaching high accuracy over a long forecasting horizon. For instance, the CERN facility was shut down most of the year for maintenance which shows no transfers or very few patterns are picked up to and from the facility. Additionally, sites such as DENV, BNL, and NASH run performance tests which consist of very small transfers that take up less bandwidth. These transfers are so small that the spatial-temporal patterns are not highly correlated. Therefore, they are not amenable for modeling and learning with \ddcrnn.  



\section{Conclusion and Future work}

We developed \ddcrnn, a dynamic diffusion convolution recurrent neural network  for forecasting traffic bandwidth in a highly dynamic research wide area network. These traffic traces show dynamic traffic and lack long term periodic patterns. Our approach is built on a diffusion convolution recurrent neural network that models spatial and temporal patterns using graph diffusion convolution operations within recurrent units. In our approach, the dynamic traffic is explicitly captured using a state-specific adjacency matrix computed from the current traffic state in the network. 


We evaluated our approach on real traffic traces from ESnet, a US research network. Our approach outperformed several other alternative methods, first, showing that the proposed \ddcrnn achieves higher forecasting accuracy than that of the static variant. And secondly, it shows that a single-model can be used to achieve higher forecasting accuracy than the site-specific models obtained from linear regression, ARIMA, random forest, gradient boosting, and simple neural network variants that were previously used for traffic forecasting.

Achieving higher prediction accuracy presents several advantages to performing informed routing decisions for potentially new flows that could be impacted with regular R-WAN behavior. Our \ddcrnn approach exposes many potentials to allow informed flow and routing allocations for network operations, such as scheduling long-running flow on alternative routes to prevent congestion points for other smaller flows. 
In the future, we will couple these decisions with a controller to test how congestion-free routing will impact the average utilization of a network in practice.

\section*{Acknowledgements}

This work was supported by the U.S. Department of Energy, Office of Science Early Career Research Program for `Large-scale Deep Learning for Intelligent Networks' Contract no  FP00006145. This material is based upon work supported by the U.S.\ Department of Energy (DOE), Office of Science, Office of Advanced Scientific Computing Research, under Contract DE-AC02-06CH11357.  This research used resources of the Argonne Leadership Computing Facility, which is a DOE Office of Science User Facility.

\bibliographystyle{IEEEtran}
\bibliography{reference}

\footnotesize{
\begin{center}
    \framebox{\parbox{3in}{
    The submitted manuscript has been created by UChicago Argonne, LLC, Operator of Argonne National Laboratory (``Argonne''). Argonne, a U.S. Department of Energy Office of Science laboratory, is operated under Contract No. DE-AC02-06CH11357.
    It is also supported by the Lawrence Berkeley National Laboratory, a U.S. Department of Energy, Office of Science Early Career Research Program for ‘Large-scale  Deep  Learning  for  Intelligent  Networks’  Contract  no. FP00006145. The U.S. Government retains for itself, and others acting on its behalf, a paid-up nonexclusive, irrevocable worldwide license in said article to reproduce, prepare derivative works, distribute copies to the public, and perform publicly and display publicly, by or on behalf of the Government. The Department of Energy will provide public access to these results of federally sponsored research in accordance with the DOE Public Access Plan. http://energy.gov/downloads/doe-public-access-plan}}
\end{center}
}

\end{document}